%% file: main.tex
\documentclass[journal,twocolumn]{IEEEtran}

\usepackage{graphicx}
\usepackage{amsmath}
\usepackage{amsfonts}
\usepackage{amssymb}
\usepackage[amssymb]{SIunits}
\usepackage{longtable}

\usepackage{import}
\usepackage{color}
\usepackage{algorithm,algpseudocode}
\usepackage{amsthm}
\usepackage{cite}
\usepackage{subcaption}

\ifCLASSINFOpdf
\else
\fi

\usepackage{commath}
\usepackage{lipsum}
\usepackage{multicol}
\usepackage{multirow}

\input{Header}

\begin{document}
\title{Predictive Analysis of COVID-19 Time-series Data from Johns Hopkins University}

\author{\IEEEauthorblockN{Alireza M. Javid, Xinyue Liang, Arun Venkitaraman, and Saikat Chatterjee} \\        \IEEEauthorblockA{\textit{School of Electrical Engineering and Computer Science} \\
\textit{KTH Royal Institute of Technology, Sweden}\\
            \{almj, xinyuel, arunv, sach\}@kth.se}
    }

\markboth{}%
{Shell \MakeLowercase{\textit{et al.}}: Bare Demo of IEEEtran.cls for IEEE Journals}

\maketitle

\begin{abstract}
We provide a predictive analysis of the spread of COVID-19, also known as SARS-CoV-2, using the dataset made publicly available online by the Johns Hopkins University. Our main objective is to provide predictions of the number of infected people for different countries in the next 14 days. The predictive analysis is done using time-series data transformed on a logarithmic scale. We use two well-known methods for prediction: polynomial regression and neural network. As the number of training data for each country is limited, we use a single-layer neural network called the extreme learning machine (ELM) to avoid over-fitting. Due to the non-stationary nature of the time-series, a sliding window approach is used to provide a more accurate prediction.
\end{abstract} 


\section{Goal}
The COVID-19 pandemic has led to a massive global crisis, caused by the rapid spread rate and severe fatality, especially, among those with a weak immune system. In this work, we use the available COVID-19 time-series of the infected cases to build models for predicting the number of cases in the near future. In particular, given the time-series till a particular day, we make predictions for the number of cases in the next $\tau$ days, where $\tau \in \{1,3,7,14\}$. This means that we predict for the next day, after 3 days, after 7 days, and after 14 days. Our analysis is based on the time-series data made publicly available on the COVID-19 Dashboard by the Center for Systems Science and Engineering (CSSE) at the Johns Hopkins University (JHU) (https://systems.jhu.edu/research/public-health/ncov/) \cite{JHU_article}.

Let $y_n$ denote the number of confirmed cases on the $n$-th day of the time-series after start of the outbreak. Then, we have the following
\begin{itemize}
    \item The input consists of the last $n$ samples of the time-series given by $\mathbf{y}_n\triangleq[y_{1}, y_{2},\cdots, y_n]$.
    \item The predicted output is $t_n=\hat{y}_{n+\tau}$, $\tau \in \{1,3,7,14\}$.
    \item Due to non-stationary nature of the time-series data, a sliding window of size $w$ is used over $\mathbf{y}_n$ to make the prediction, and $w$ is found via cross-validation.
    \item The predictive function $f(\cdot)$ is modeled either by a polynomial or a neural network, and is used to make the prediction:
    \begin{equation*}
        \boxed{\hat{y}_{n+\tau}=f(\mathbf{y}_n)}
    \end{equation*}
\end{itemize}

\begin{table}[t!]
    \centering
    \caption{Countries considered from JHU dataset.}
    \label{table:Countries}
    \begin{tabular}{|c|}
    \hline
    Countries considered in the analysis \\
    \hline
    \hline
    Sweden\\
    Denmark\\
    Finland\\
    Norway\\
    France\\
    Italy\\
    Spain\\
    UK\\
    China\\
    India\\
    Iran\\
    USA\\
    \hline
    \end{tabular}
\end{table}

\section{Dataset}
The dataset from JHU contains the cumulative number of cases reported daily for different countries. We base our analysis on 12 of the countries listed in Table \ref{table:Countries}. For each country, we consider the time-series $\mathbf{y}_n$ starting from the day when the first case was reported. Given the current day index $n$, we predict the number of cases for the day $n+\tau$ by considering as input the number of cases reported for the past $w$ days, that is, for the days $n-w+1$ to $n$.

\section{Approaches}

We use data-driven prediction approaches without considering any other aspect, for example, models of infectious disease spread \cite{folkhal}.
We apply two approaches to analyze the data to make predictions, or in other words, to learn the function $f$:
\begin{itemize}
    \item{\it  Polynomial model approach}: Simplest curve fit or approximation model, where the number of cases is approximated locally with polynomials $-$ $f$ is a polynomial.
    \item{\it Neural network approach}: A supervised learning approach that uses training data in the form of input-output pairs to learn a predictive model $-$ $f$ is a neural network.
\end{itemize}
We describe each approach in detail in the following subsections.

\subsection{Polynomial model}
\subsubsection{Model} We model the expected value of $y_n$ as a third degree polynomial function of the day number $n$:
\begin{equation*}
    \boxed{f(n)=p_0 + p_1 n^1 + p_2 n^2 + p_3 n^3}
\end{equation*}
The set of coefficients $\{p_0,p_1,p_2,p_3\}$ are learned using the available training data. Given the highly non-stationary nature of the time-series, we consider local polynomial approximations of the signal over a window of $w$ days, instead of using all the data to estimate a single polynomial $f(\cdot)$ for the entire time-series. Thus, at the $n$-th day, we learn the corresponding polynomial $f(\cdot)$ using $\mathbf{y}_{n,w}\triangleq[y_{n-w+1},\cdots, y_{n-1}, y_n]$.  
\subsubsection{How the model is used} 
Once the polynomial is determined, we use it to predict for ${(n+\tau)}$-th day as 
\begin{equation*}
    \boxed{\hat{y}_{n+\tau}=f(n+\tau)}
\end{equation*}
For every polynomial regression model, we construct the corresponding polynomial function $f(\cdot)$ by using $\mathbf{y}_{n,w}$ as the most recent input data of size $w$. 
The appropriate window size $w$ is found through cross-validation.

\subsection{Neural networks}

\subsubsection{Model}
We use Extreme Learning Machine (ELM) as the neural network model to avoid overfitting to the training data. As the length of the time-series data for each country is limited, the number of training samples for the neural network would be quite small, which can lead to severe overfitting in large scale neural network such as deep neural networks (DNNs), convolutional neural networks (CNNs), etc. \cite{DNN_2013,CNN_2012}. ELM, on the other hand, is a single layer neural network which uses random weights in its first hidden layer \cite{elm_HUANG2015}. The use of random weights has gained popularity due to its simplicity and effectiveness in training \cite{giryes_randomweights,SSFN_2020,HNF_2020}.  We now briefly describe ELM.

Consider a dataset containing $N$ samples of pair-wise $P$-dimensional input data $\mathbf{x} \in \mathbb{R}^{P}$ and the corresponding $Q$-dimensional target vector $\mathbf{t} \in \mathbb{R}^{Q}$ as $\mathcal{D}=\{(\mathbf{x}_n,\mathbf{t}_n)\}_{n=1}^N$. We construct the feature vector as $\mathbf{z}_n = \mathbf{g}(\mathbf{W} \mathbf{x}_n) \in \mathbb{R}^h$, where 
\begin{itemize}
    \item weight matrix $\mathbf{W} \in \mathbb{R}^{h \times P}$ is an instance of Normal distribution, 
    \item $h$ is the number of hidden neurons, and
    \item activation function $\mathbf{g}(\cdot)$ is the rectified linear unit (ReLU).
\end{itemize}
To predict the target, we use a linear projection of feature vector $\mathbf{z}_n$ onto the target. Let the predicted target for the $n$-th sample be $\mathbf{O} \mathbf{z}_n$. Note that $\mathbf{O} \in \mathbb{R}^{Q \times h}$. By using $\ell_2$-norm regularization, we find the optimal solution for the following convex optimization problem
\begin{flalign}
\mathbf{O}^{\star} & = \arg\displaystyle\min_{\mathbf{O}} \sum_{n=1}^{N} \| \mathbf{t}_n - \mathbf{O} \mathbf{z}_n \|^2_2  + \lambda \|\mathbf{O} \|_F^2, \label{eq:Optimization_VW}
\end{flalign}
where $\| \cdot \|_F$ denotes the Frobenius norm. Once the matrix $\mathbf{O}^{\star}$ is learned, the prediction for any new input $\mathbf{x}$ is now given by
\begin{equation*}
   \boxed{ \hat{\mathbf{t}}=\mathbf{O}^{\star} \mathbf{g}(\mathbf{W}\mathbf{x})}
\end{equation*}
\subsubsection{How the model is used}
When using ELM to predict the number of cases, we define $\mathbf{x}_n=[y_{n-w+1}, ... , y_{n-1}, y_{n}]^{\top}$ and $\mathbf{t}_n=[y_{n+\tau}]$. Note that $\mathbf{x}_n \in \mathbb{R}^w$ and $\mathbf{t}_n \in \mathbb{R}$. For a fixed $\tau \in \{1,3,7,14\}$, we use cross-validation to find the proper window size $w$, number of hidden neurons $h$, and the regularization hyperparameter $\lambda$.

\section{Experiments}

\subsection{With the available data till May 4, 2020}
\label{subSec:May4}
In this subsection, we make predictions based on the time-series data which currently is available until today May 4, 2020, for $\tau \in \{1,3,7\}$.
We estimate the number of cases for the last 31 days of the countries in Table \ref{table:Countries}. For each value of $\tau \in \{1,3,7\}$, we compare the estimated number of cases $\hat{y}_{n+\tau}$ with the true value $y_{n+\tau}$ and report the estimation error in percentage, i.e., 
\begin{align}
\text{Error} = \frac{|y_{n+\tau}-\hat{y}_{n+\tau}|}{y_{n+\tau}} \times 100.
\end{align}
We carry out two sets of experiments for each of the two approaches (polynomial and ELM) to examine their sensitivity to the new arriving training samples. In the first set of experiments, we implement cross-validation to find the hyperparameters without using the observed samples of the time-series as we proceed through 31 days span. In the second set of experiments, we implement cross-validation in a daily manner as we observe new samples of the time-series. In the latter setup, the window size $w$ varied with respect to time to find the optimal hyperparameters as we proceed through time. We refer to this setup as 'ELM time-varying' and 'Poly time-varying' in the rest of the manuscript. 

\begin{figure*}[t!]
    \begin{subfigure}{\textwidth}
        \centering
        \includegraphics[width=\textwidth]{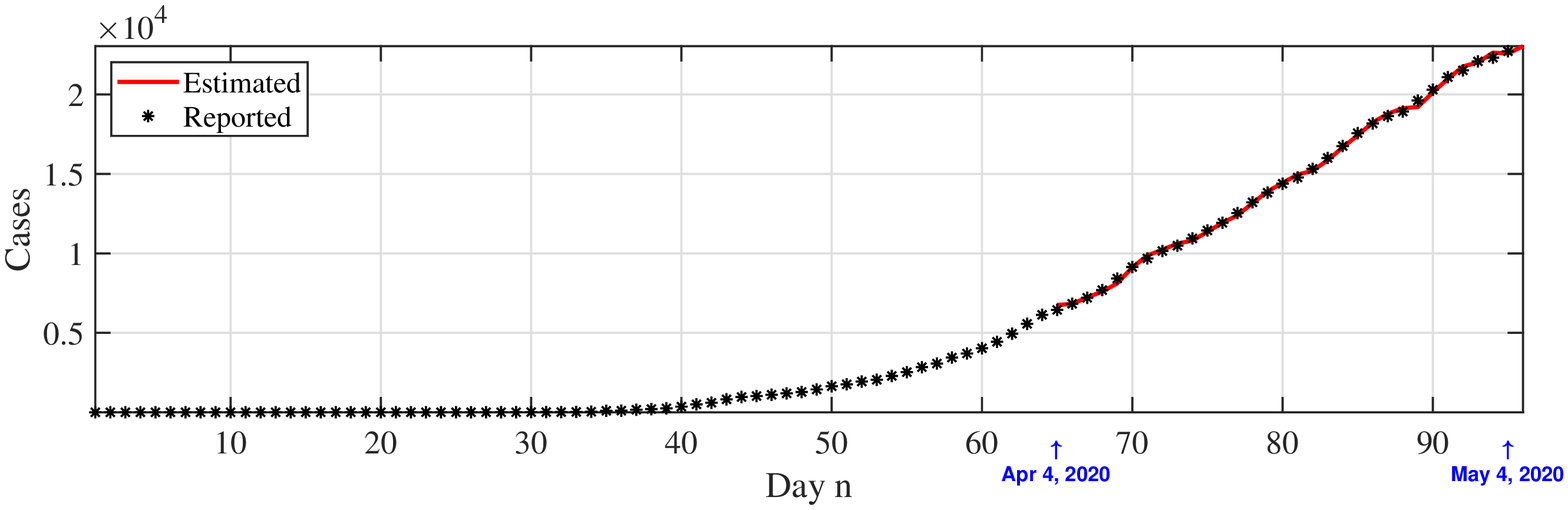}
        \caption{$\tau=1$}
    \end{subfigure} \\
    \begin{subfigure}{\textwidth}
        \centering
        \includegraphics[width=\textwidth]{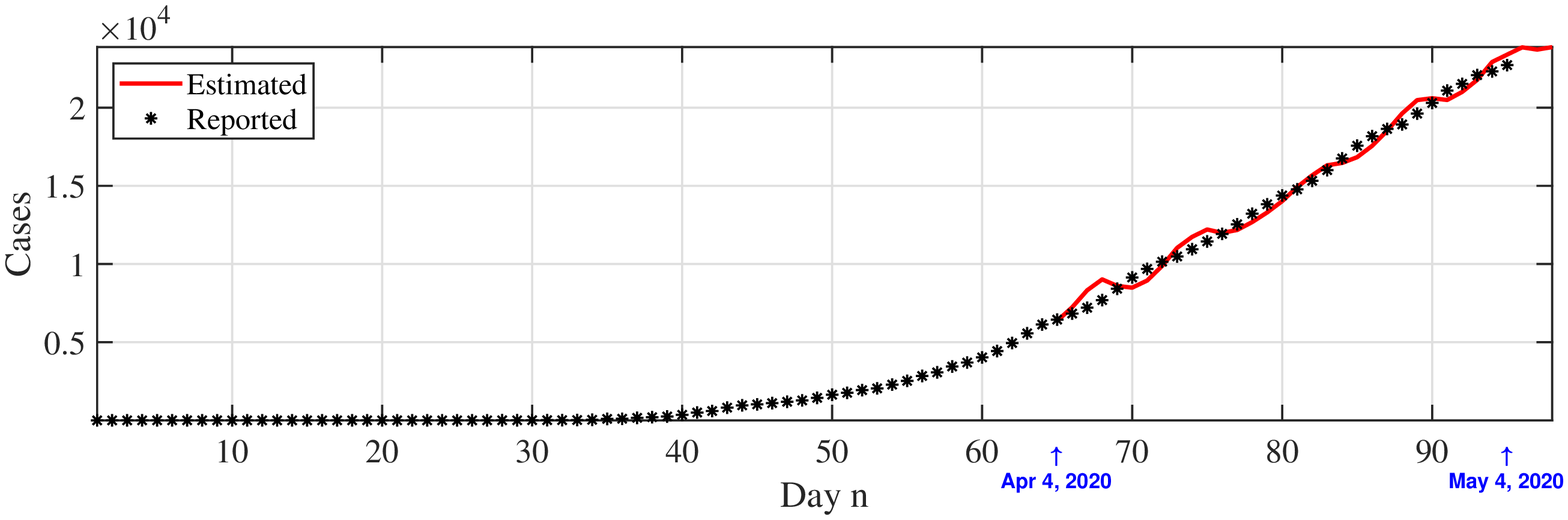}
        \caption{$\tau=3$}
    \end{subfigure} \\
    \begin{subfigure}{\textwidth}
        \centering
        \includegraphics[width=\textwidth]{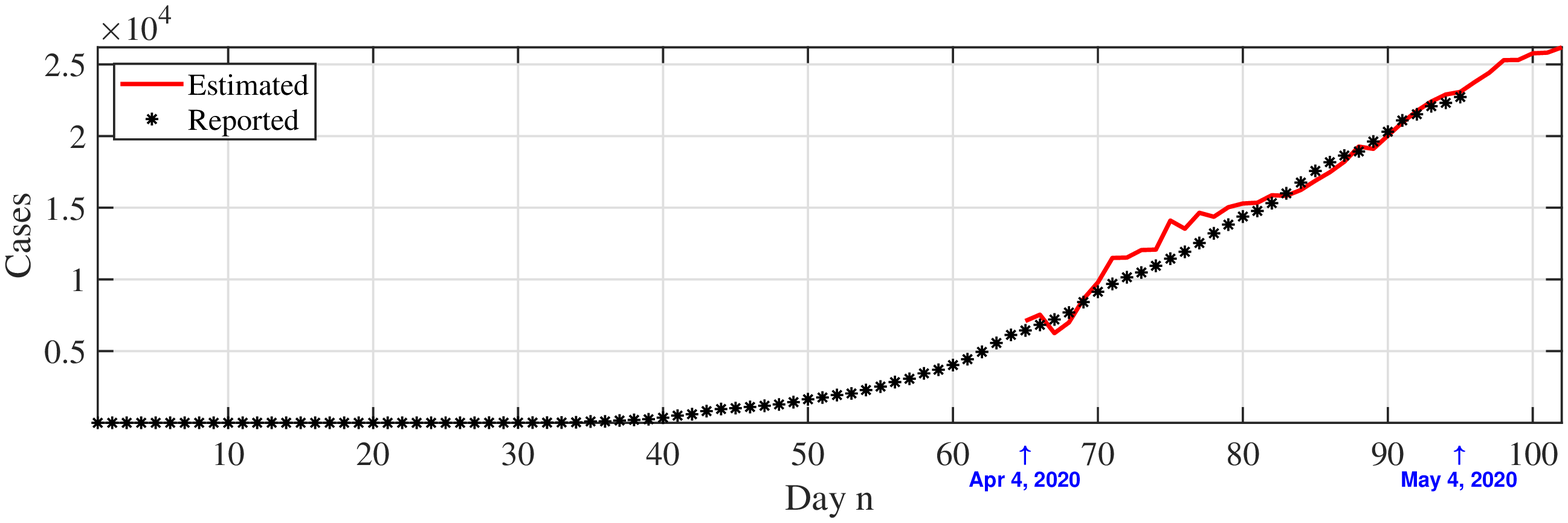}
        \caption{$\tau=7$}
    \end{subfigure} \\
    \caption{Reported and estimated cases after $\tau$ days over the last 31 days of Sweden for ELM time-varying. Update May 4.}
    \label{fig:Sweden_cases}
\end{figure*}

We first show the reported and estimated number of infection cases for Sweden by using ELM time-varying for different $\tau$'s in Figure \ref{fig:Sweden_cases}. For each $\tau$, we estimate the number of cases up to $\tau$ days after which JHU data is collected. In our later experiments, we show that ELM time-varying is typically more accurate than the other three methods (polynomial, Poly time-varying, and ELM). This better accuracy conforms to the non-stationary behavior of the time-series data, or in other words that the best model parameters change over time. Hence, the result of ELM time-varying is shown explicitly for Sweden. According to our experimental result, we predict that a total of 23039, 23873, and 26184 people will be infected in Sweden on May 5, May 7, and May 11, 2020, respectively.

\begin{figure*}[t!]
    \begin{subfigure}{\textwidth}
        \centering
        \includegraphics[width=\textwidth]{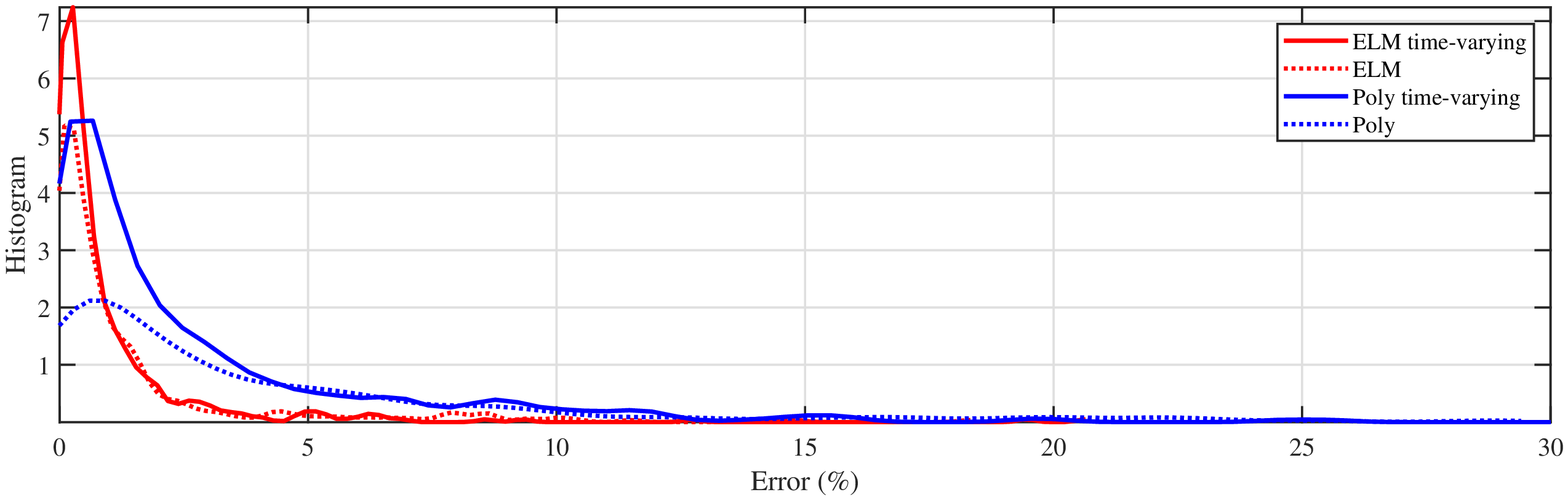}
        \caption{$\tau=1$}
        \vspace{15pt}
    \end{subfigure} \\
    \begin{subfigure}{\textwidth}
        \centering
        \includegraphics[width=\textwidth]{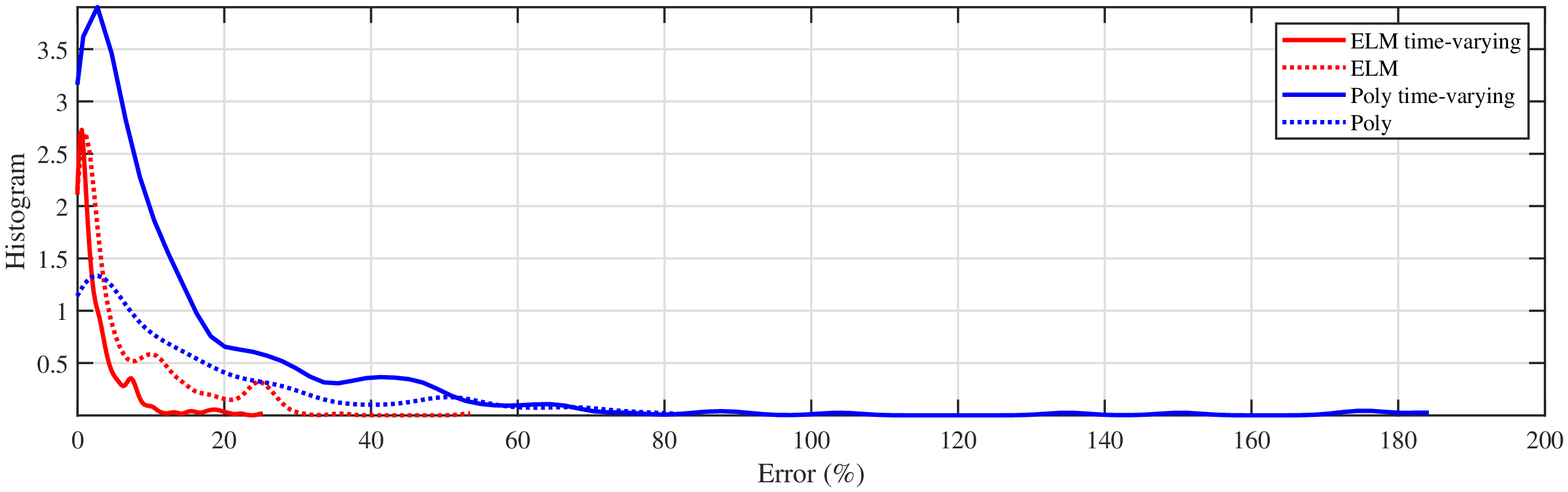}
        \caption{$\tau=3$}
        \vspace{15pt}
    \end{subfigure} \\
    \begin{subfigure}{\textwidth}
        \centering
        \includegraphics[width=\textwidth]{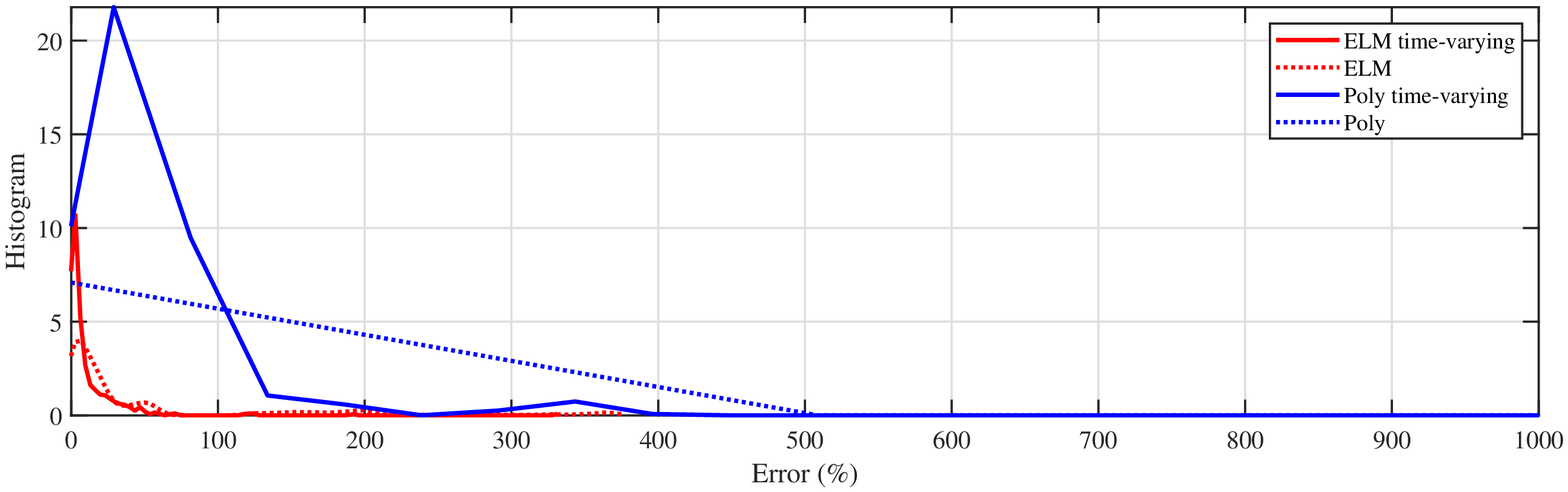}
        \caption{$\tau=7$}
        \vspace{15pt}
    \end{subfigure} \\
    \caption{Histogram of estimation error percentage over 31 days of all 12 countries for different values of $\tau$ for each of the ELM and polynomial approaches. Update May 4.}
    \label{fig:Histogram_tau}
\end{figure*}

\begin{figure*}[t!]
    \begin{subfigure}{\textwidth}
        \centering
        \includegraphics[width=\textwidth]{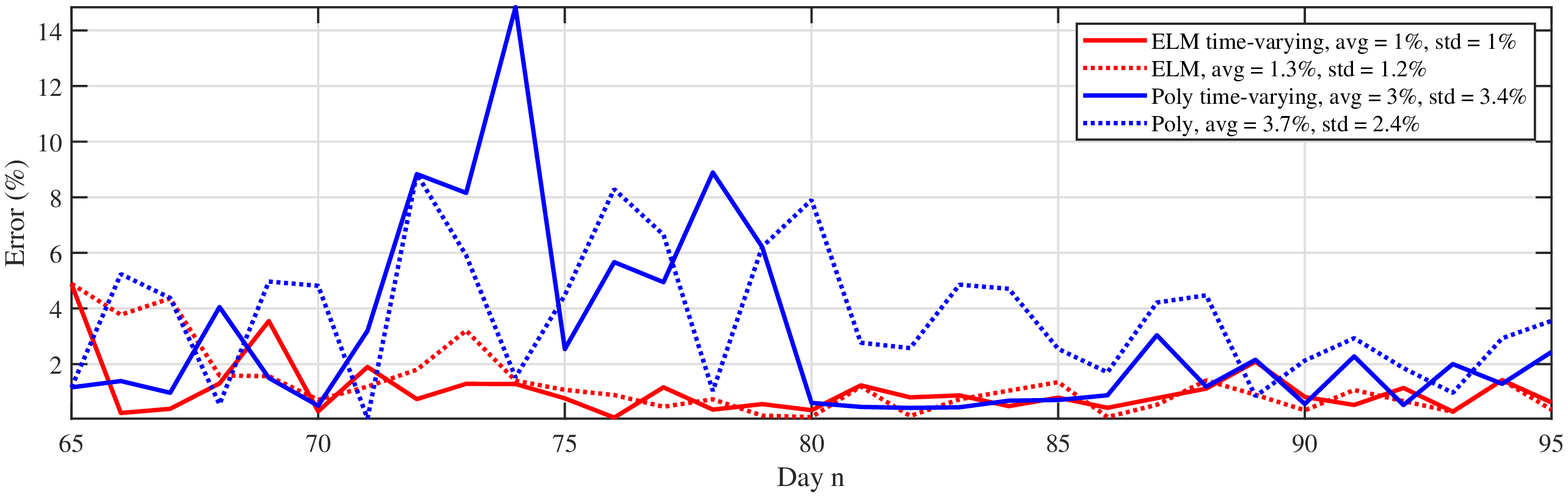}
        \caption{Sweden}
    \end{subfigure}    \\
    \begin{subfigure}{\textwidth}
        \centering
        \includegraphics[width=\textwidth]{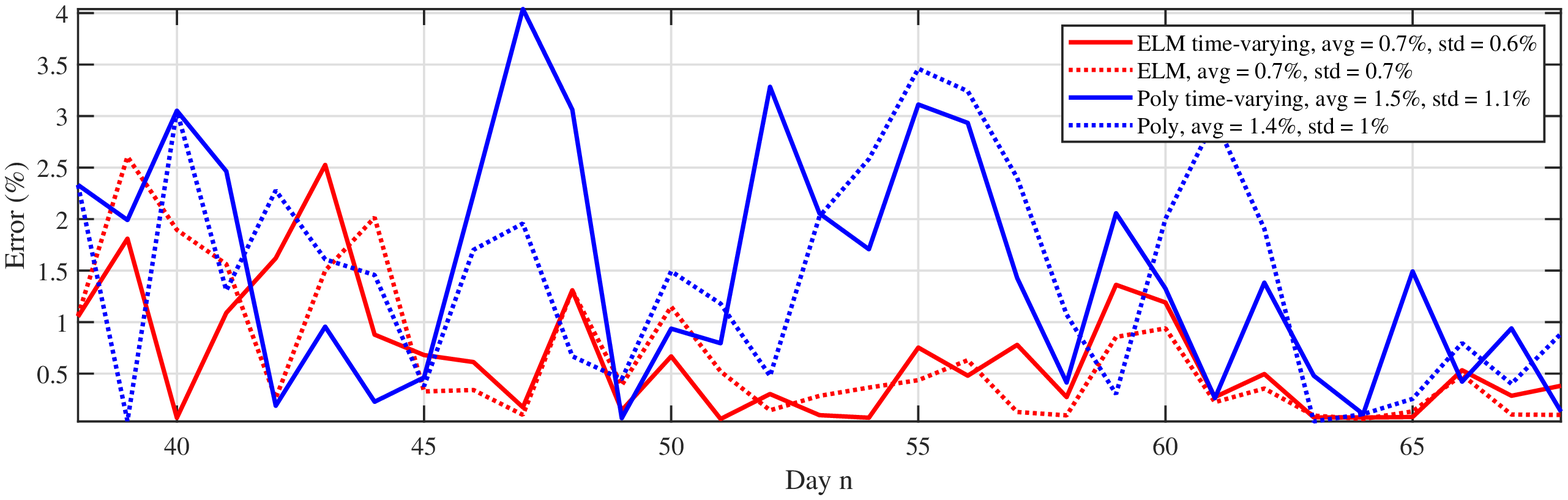}
        \caption{Denmark}
    \end{subfigure} \\
    \begin{subfigure}{\textwidth}
        \centering
        \includegraphics[width=\textwidth]{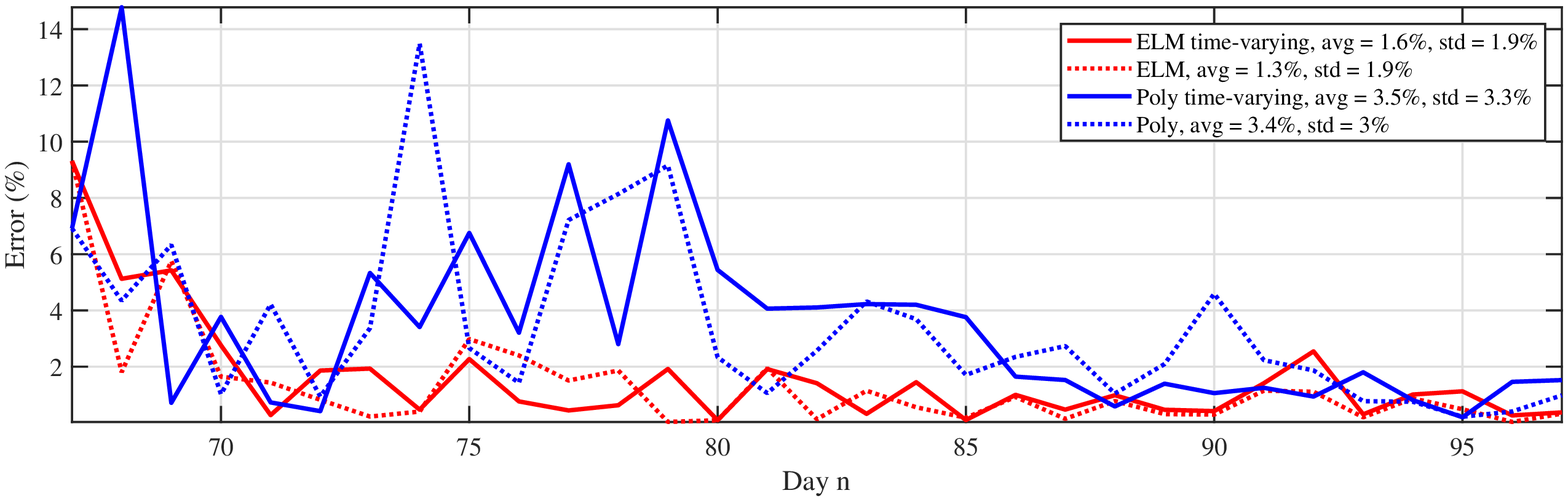}
        \caption{Finland}
    \end{subfigure}    \\
    \begin{subfigure}{\textwidth}
        \centering
        \includegraphics[width=\textwidth]{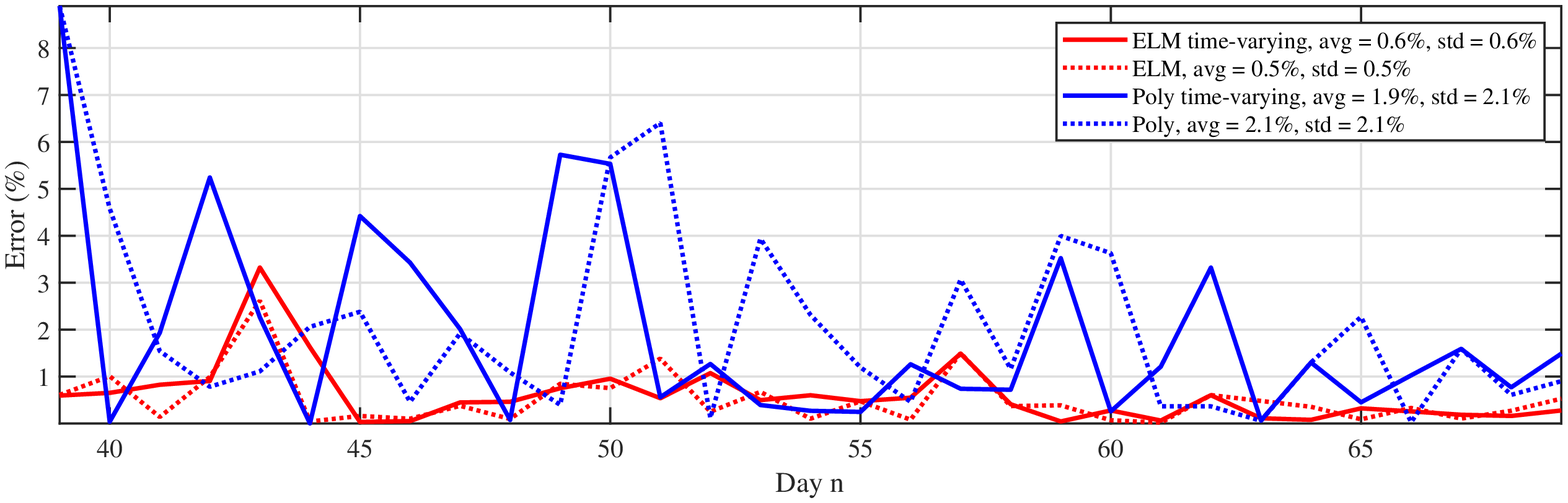} 
        \caption{Norway}
    \end{subfigure} \\
    \caption{Daily error percentage of the last 31 days of 12 countries for ELM and polynomial regression. Here, $\tau=1$.}
    \label{fig:Scandinavia_tau_1}
\end{figure*}

\begin{figure*}[t!]
    \begin{subfigure}{\textwidth}
        \centering
        \includegraphics[width=\textwidth]{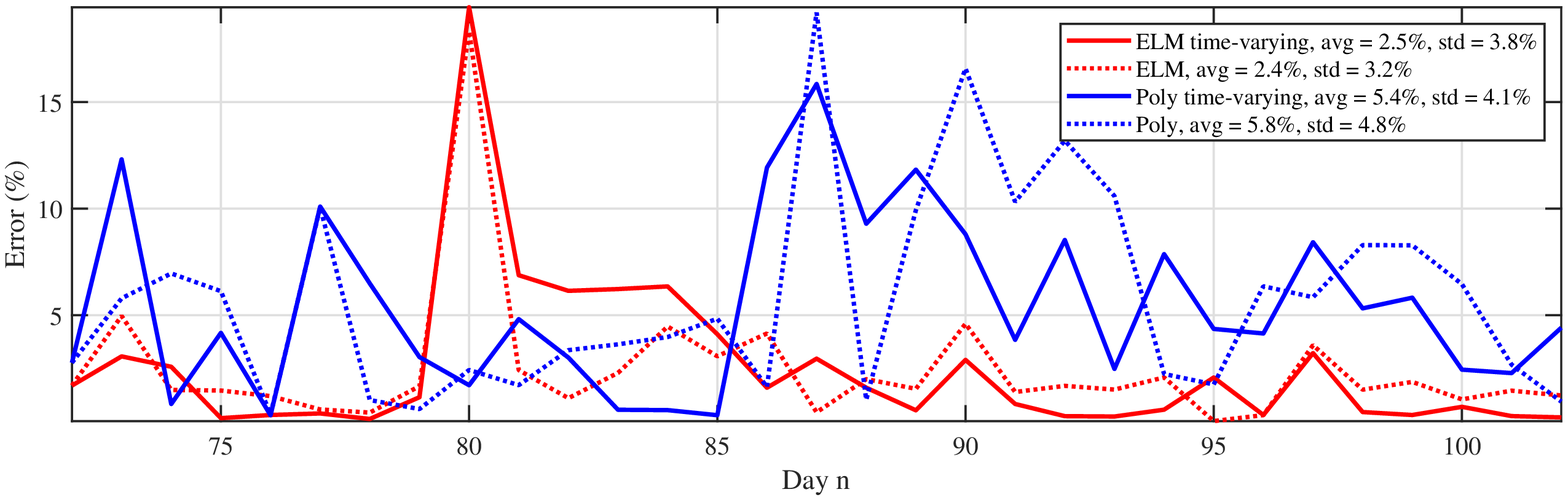} 
        \caption{France}
    \end{subfigure} \\
    \begin{subfigure}{\textwidth}
        \centering
        \includegraphics[width=\textwidth]{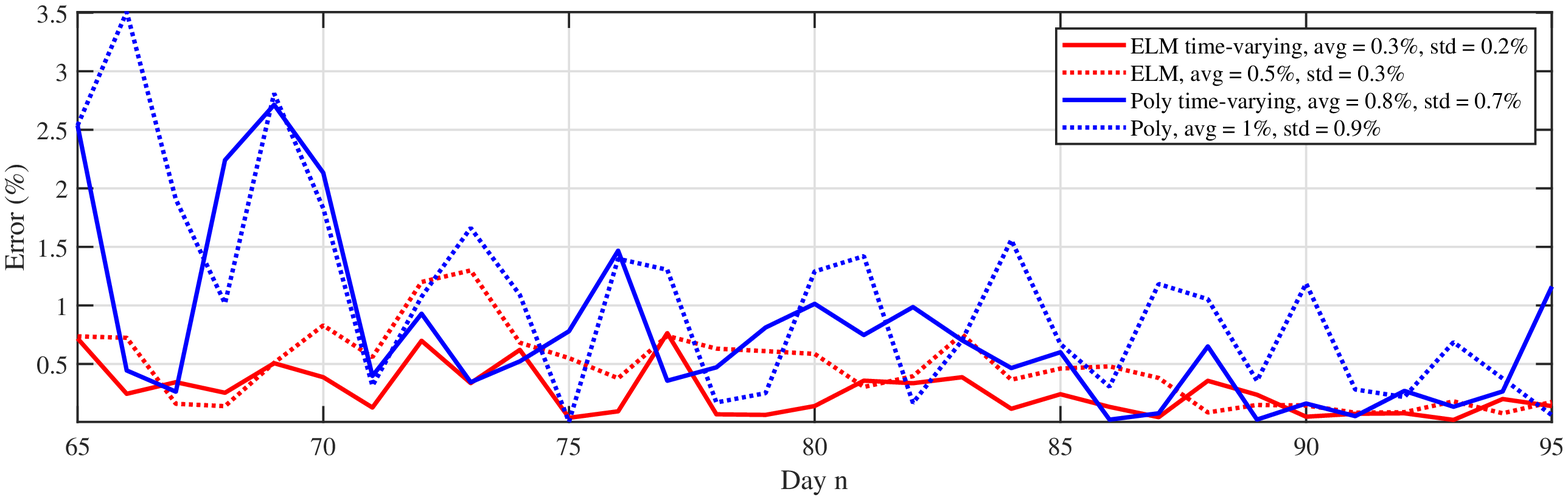}
        \caption{Italy}
    \end{subfigure}    \\
    \begin{subfigure}{\textwidth}
        \centering
        \includegraphics[width=\textwidth]{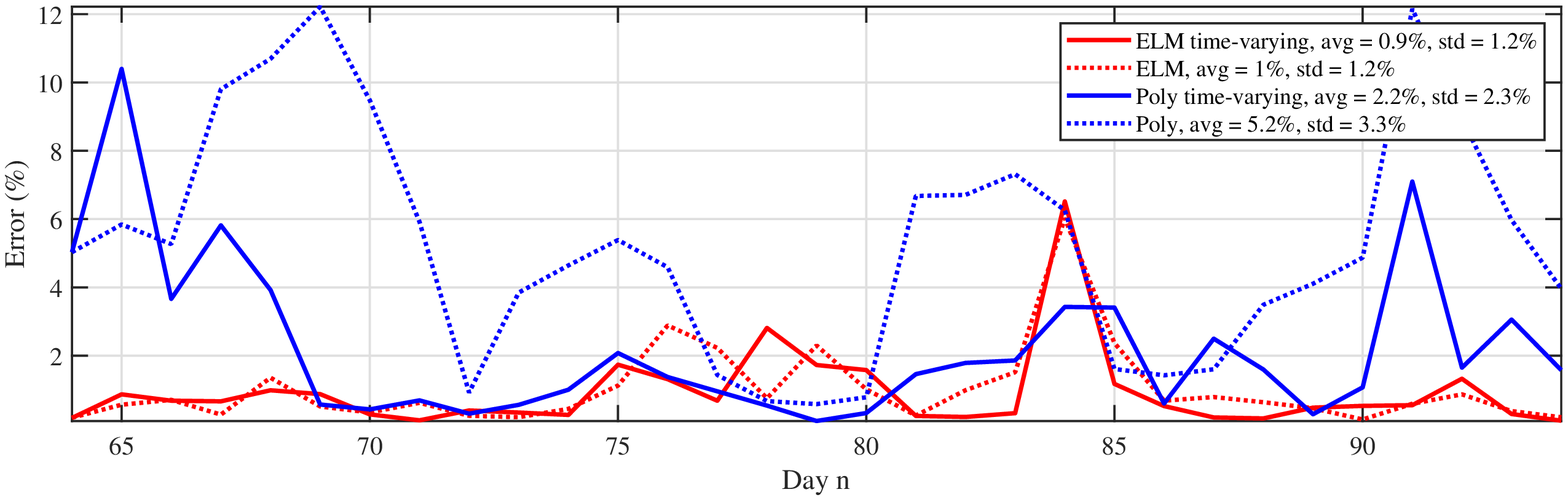}
        \caption{Spain}
    \end{subfigure} \\
    \begin{subfigure}{\textwidth}
        \centering
        \includegraphics[width=\textwidth]{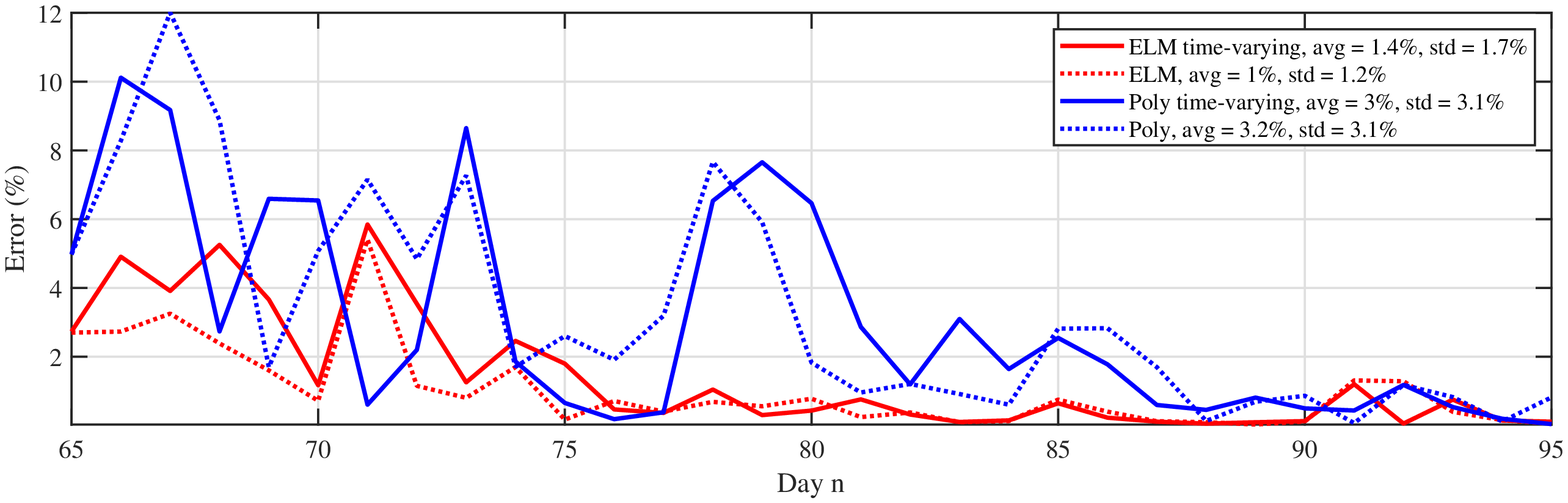} 
        \caption{UK}
    \end{subfigure} \\
    \caption{Daily error percentage of the last 31 days of 12 countries for ELM and polynomial regression. Here, $\tau=1$.}
    \label{fig:Europe_tau_1}
\end{figure*}

\begin{figure*}[t!]
    \begin{subfigure}{\textwidth}
        \centering
        \includegraphics[width=\textwidth]{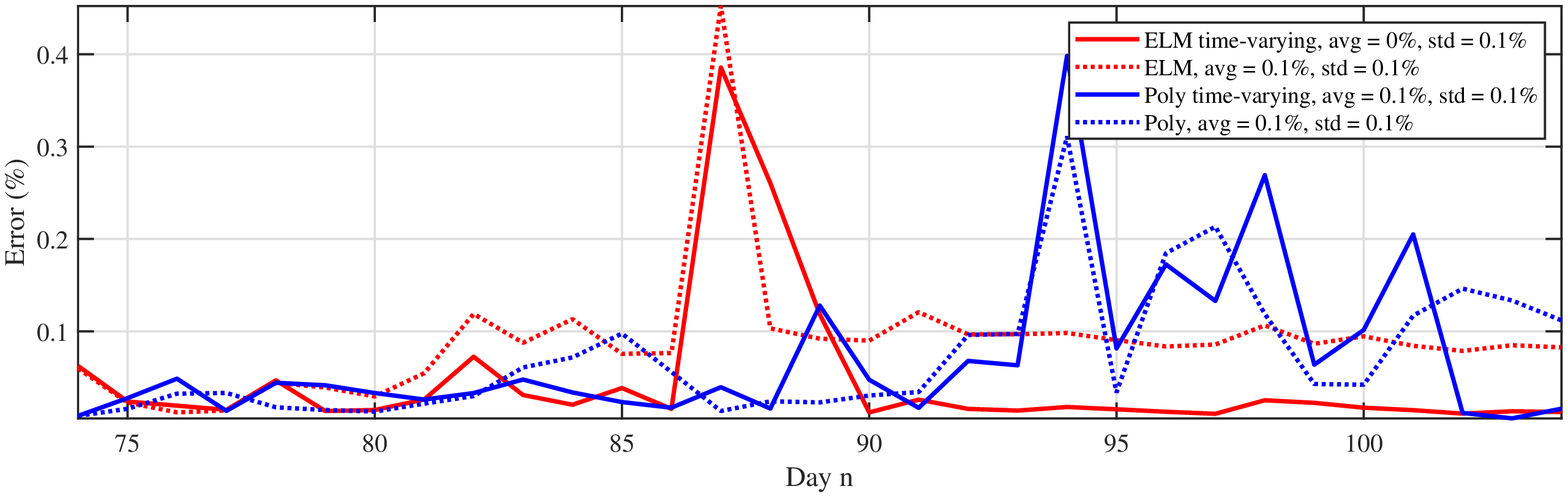} 
        \caption{China}
    \end{subfigure} \\
    \begin{subfigure}{\textwidth}
        \centering
        \includegraphics[width=\textwidth]{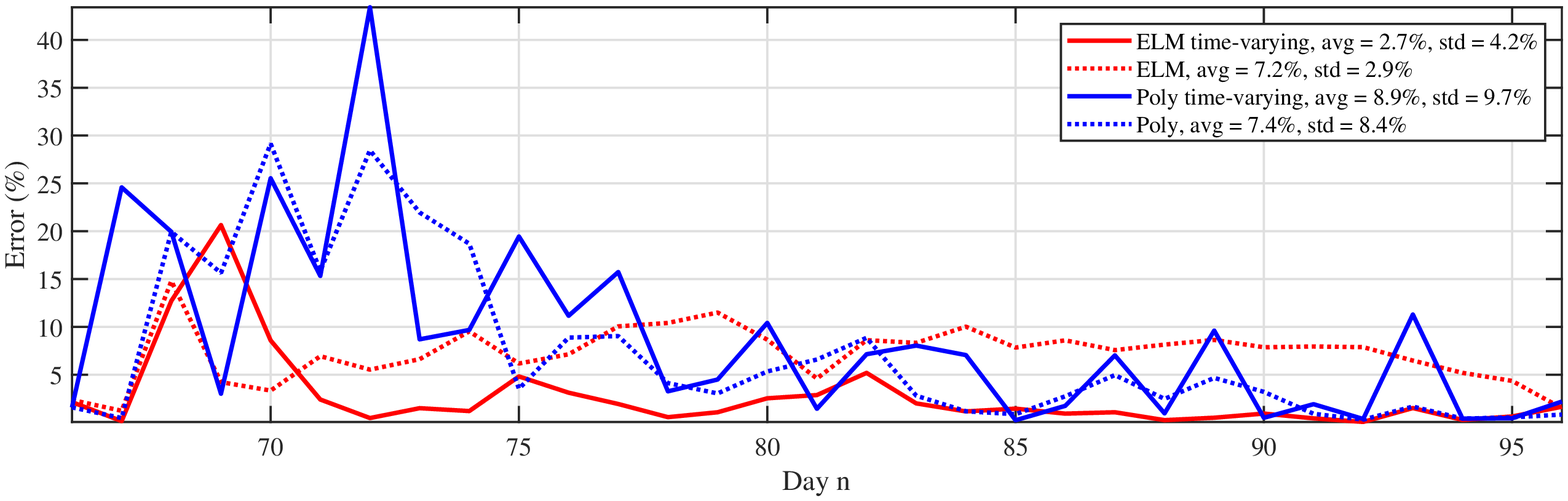} 
        \caption{India}
    \end{subfigure} \\
    \begin{subfigure}{\textwidth}
        \centering
        \includegraphics[width=\textwidth]{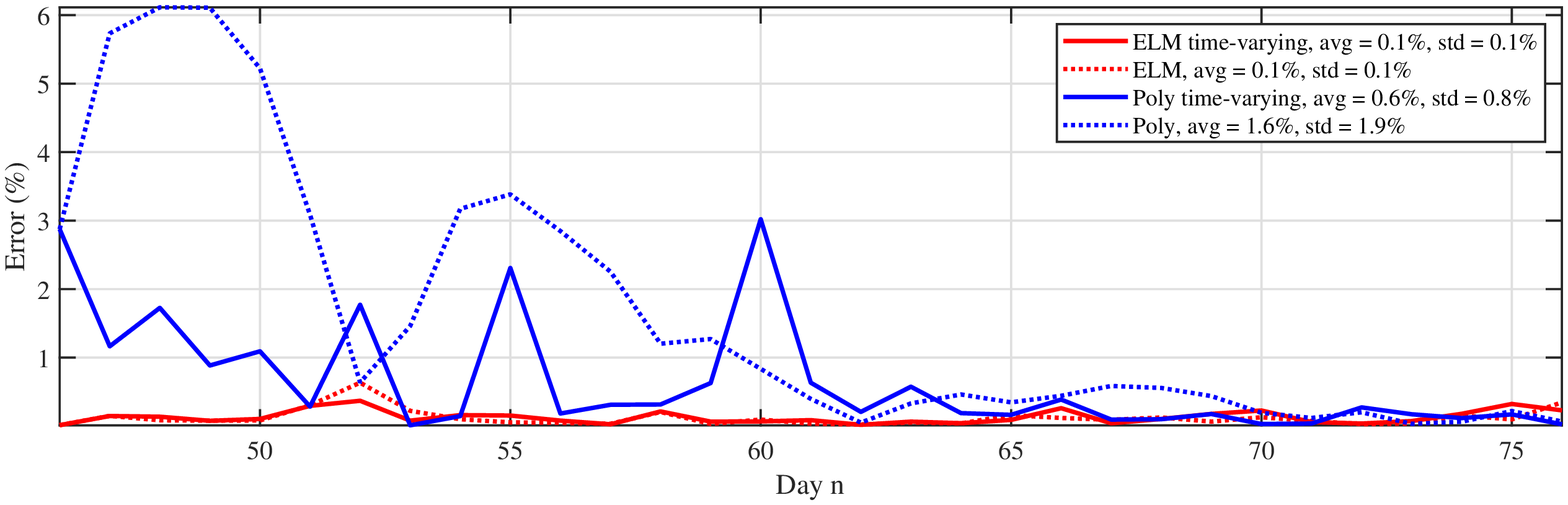}
        \caption{Iran}
    \end{subfigure} \\
    \begin{subfigure}{\textwidth}
        \centering
        \includegraphics[width=\textwidth]{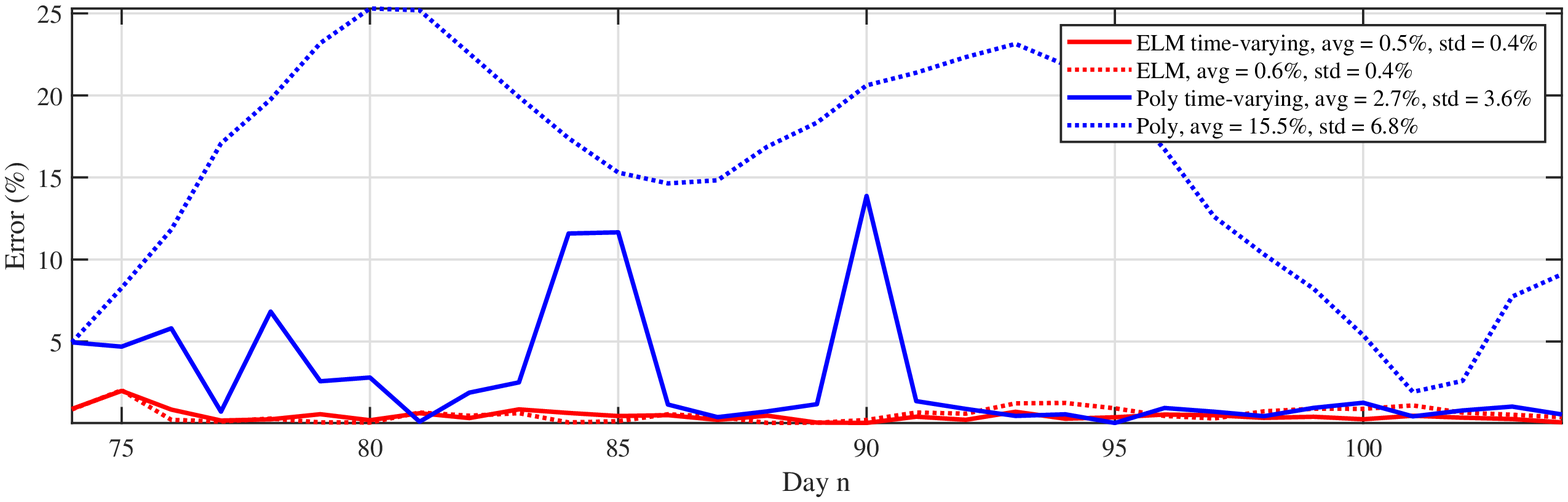} 
        \caption{USA}
    \end{subfigure} \\
    \caption{Daily error percentage of the last 31 days of 12 countries for ELM and polynomial regression. Here, $\tau=1$.}
    \label{fig:Rest_tau_1}
\end{figure*}

\begin{figure*}[t!]
    \begin{subfigure}{\textwidth}
        \centering
        \includegraphics[width=\textwidth]{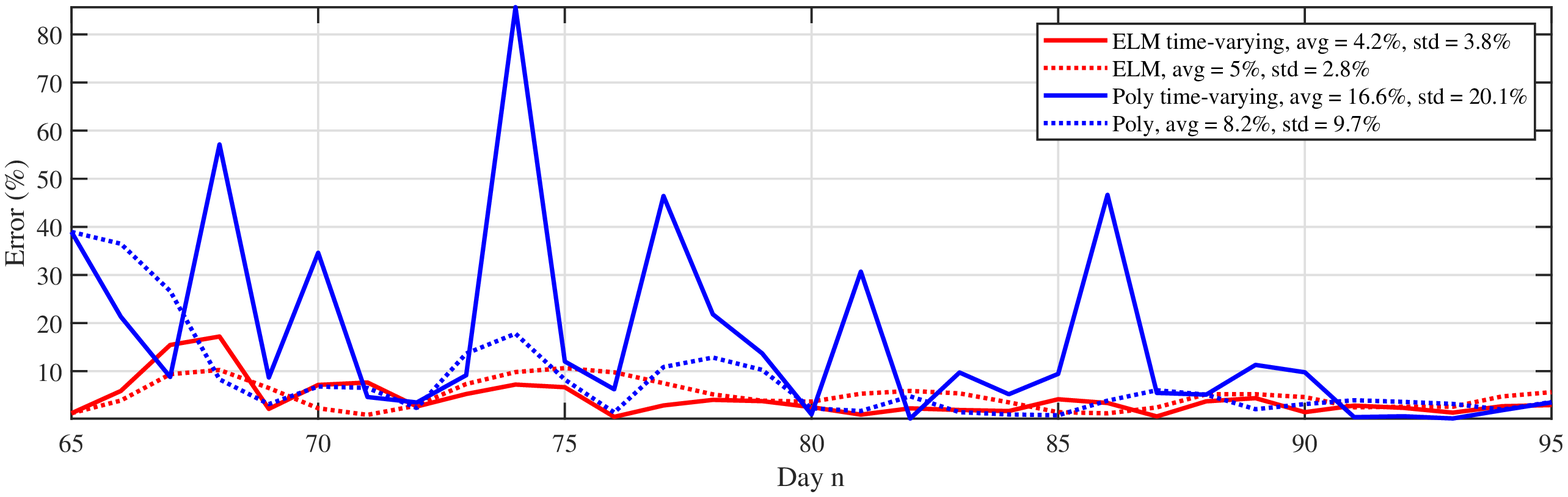}
        \caption{Sweden}
    \end{subfigure}    \\
    \begin{subfigure}{\textwidth}
        \centering
        \includegraphics[width=\textwidth]{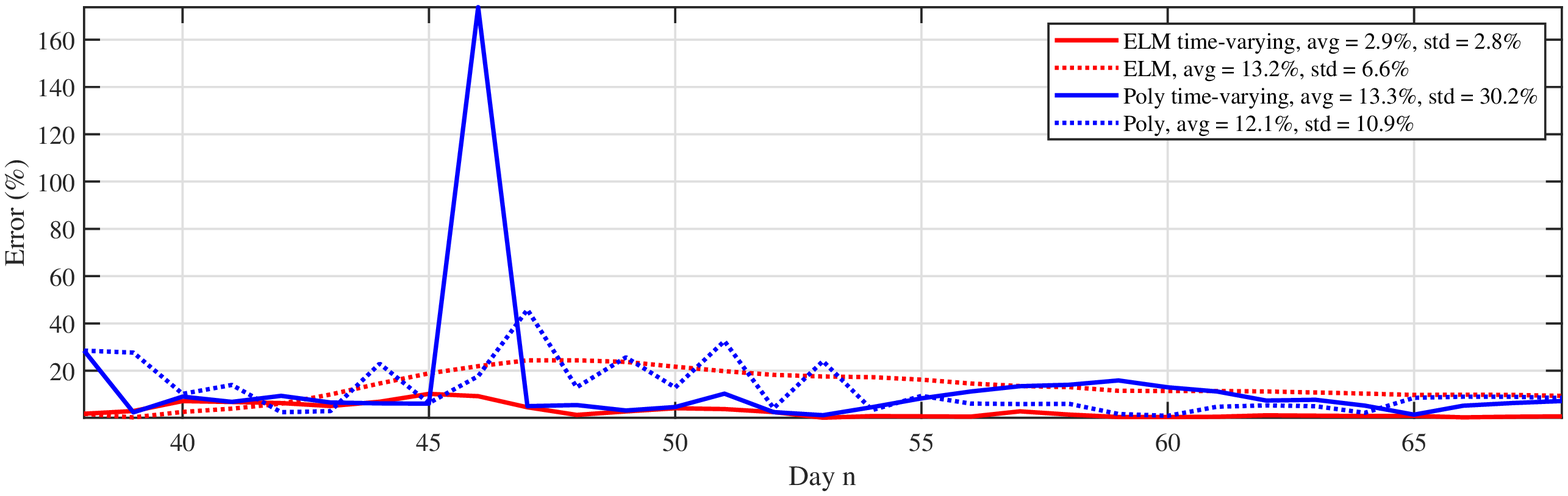}
        \caption{Denmark}
    \end{subfigure} \\
    \begin{subfigure}{\textwidth}
        \centering
        \includegraphics[width=\textwidth]{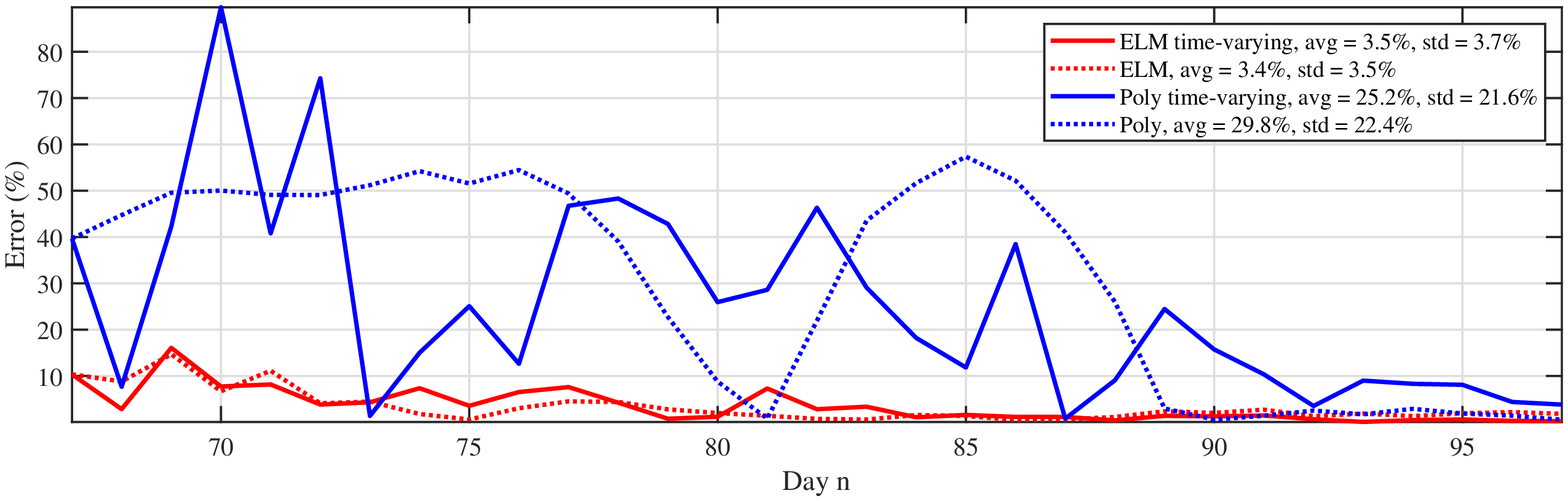}
        \caption{Finland}
    \end{subfigure}    \\
    \begin{subfigure}{\textwidth}
        \centering
        \includegraphics[width=\textwidth]{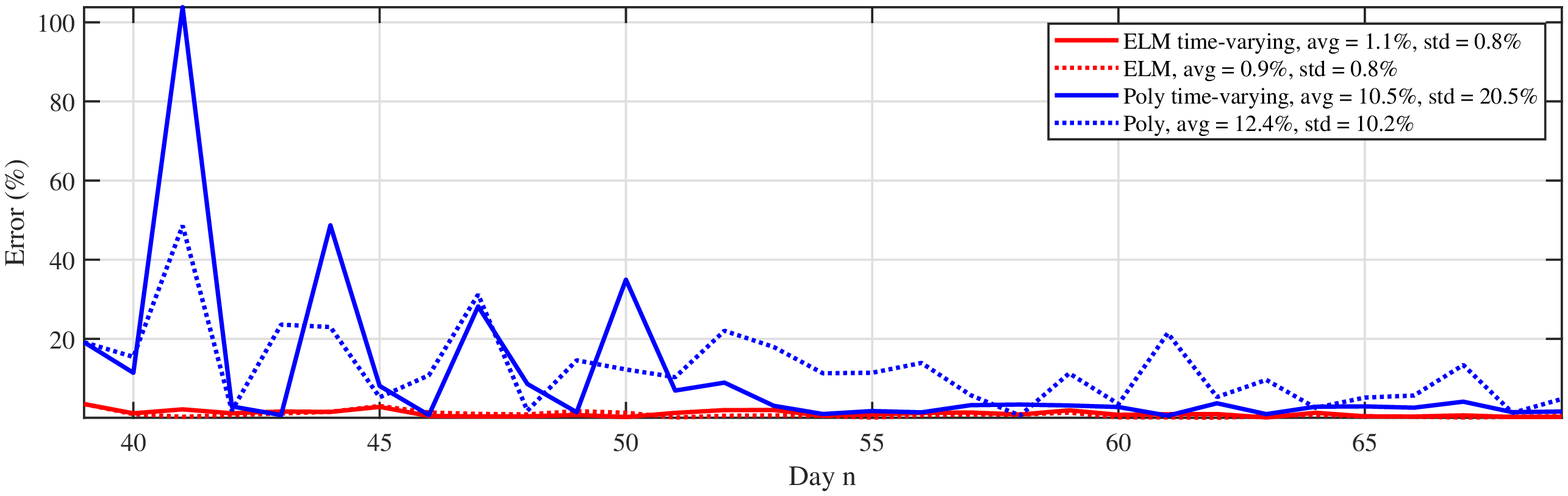} 
        \caption{Norway}
    \end{subfigure} \\
    \caption{Daily error percentage of the last 31 days of 12 countries for ELM and polynomial regression. Here, $\tau=3$.}
    \label{fig:Scandinavia_tau_3}
\end{figure*}

\begin{figure*}[t!]
    \begin{subfigure}{\textwidth}
        \centering
        \includegraphics[width=\textwidth]{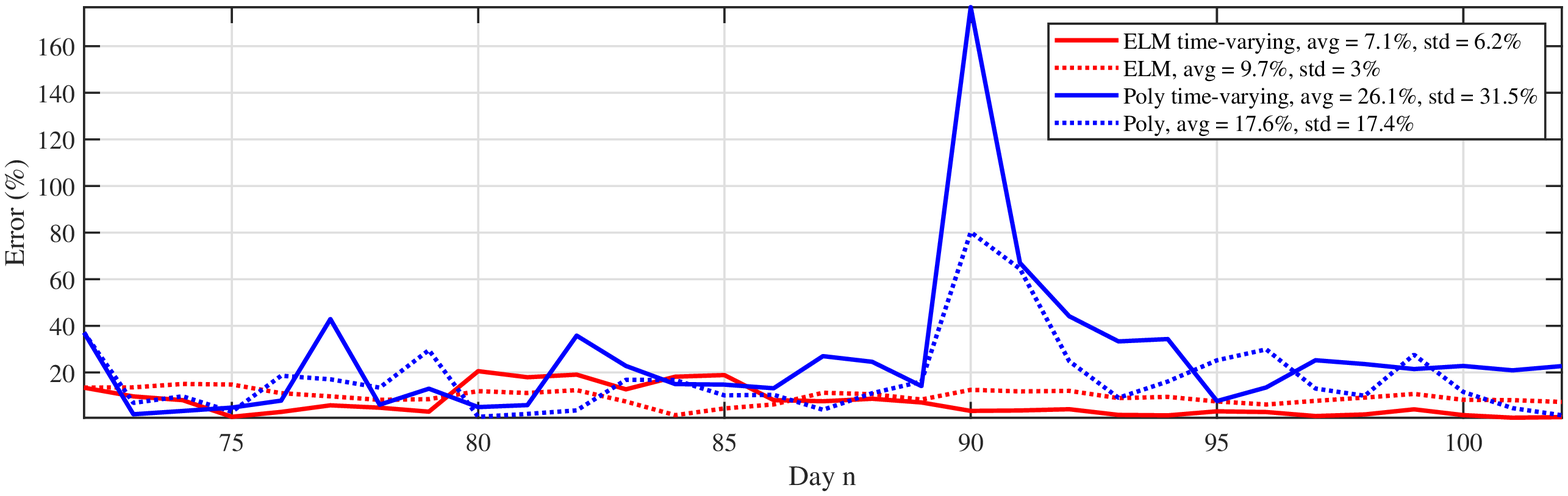} 
        \caption{France}
    \end{subfigure} \\
    \begin{subfigure}{\textwidth}
        \centering
        \includegraphics[width=\textwidth]{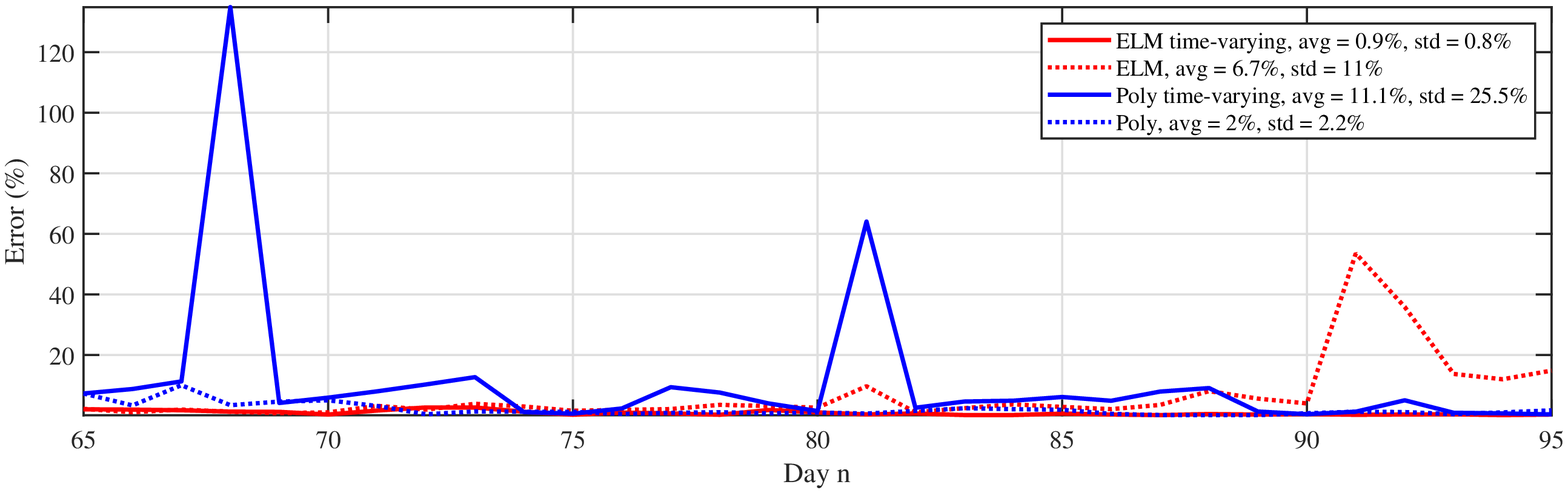}
        \caption{Italy}
    \end{subfigure}    \\
    \begin{subfigure}{\textwidth}
        \centering
        \includegraphics[width=\textwidth]{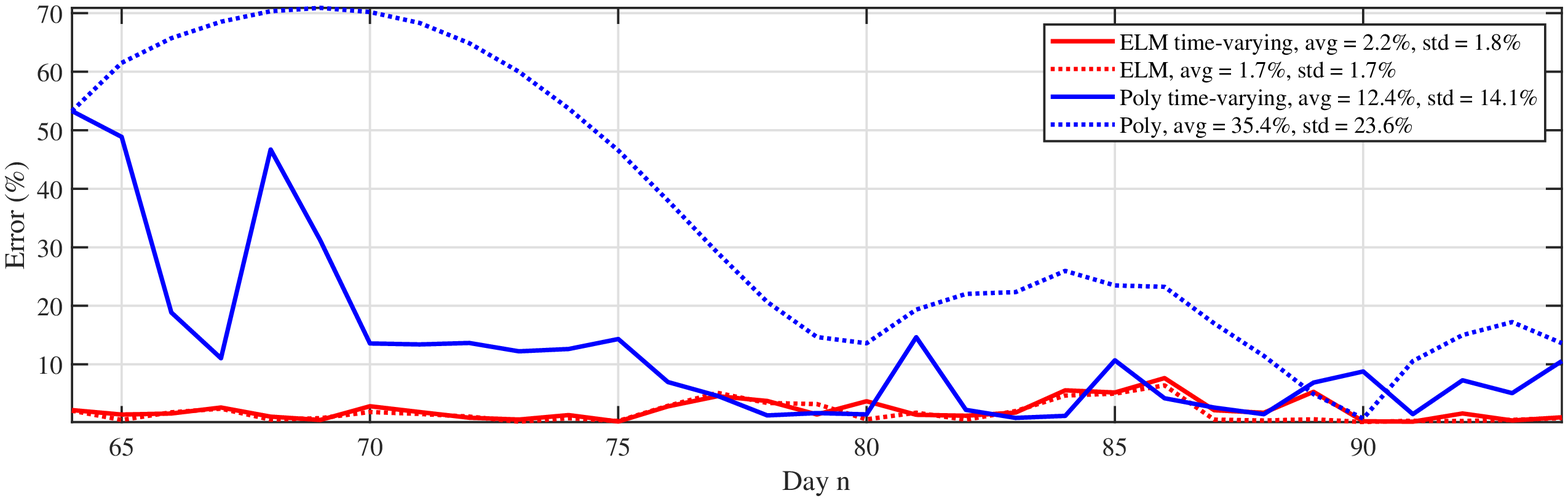}
        \caption{Spain}
    \end{subfigure} \\
    \begin{subfigure}{\textwidth}
        \centering
        \includegraphics[width=\textwidth]{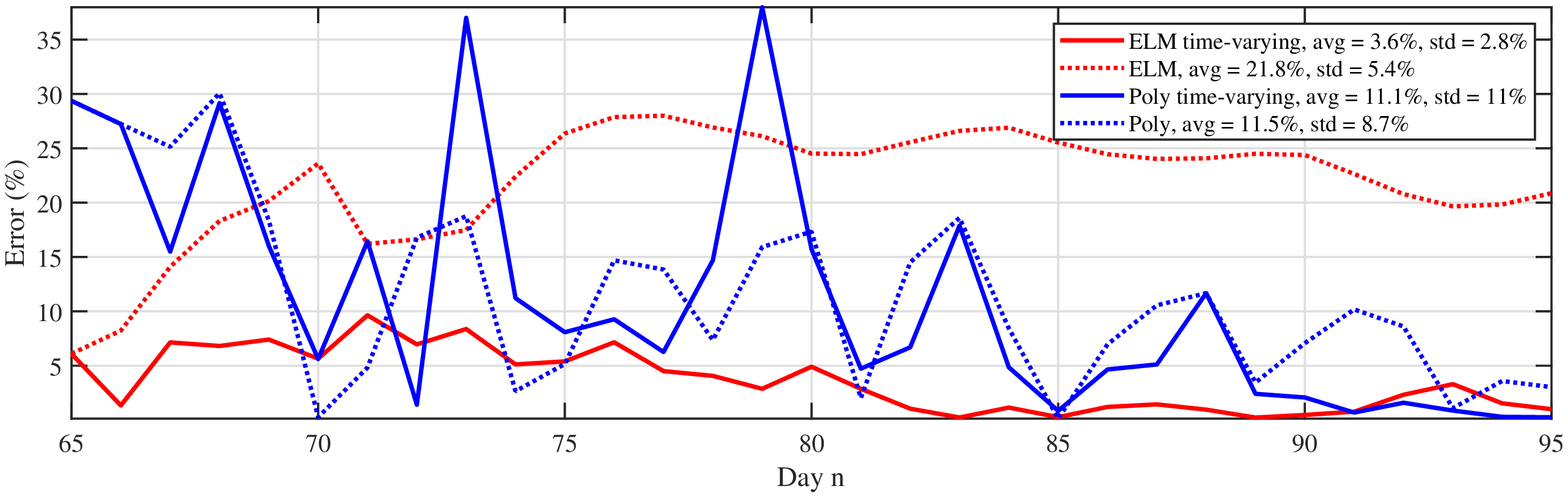} 
        \caption{UK}
    \end{subfigure} \\
    \caption{Daily error percentage of the last 31 days of 12 countries for ELM and polynomial regression. Here, $\tau=3$.}
    \label{fig:Europe_tau_3}
\end{figure*}

\begin{figure*}[t!]
    \begin{subfigure}{\textwidth}
        \centering
        \includegraphics[width=\textwidth]{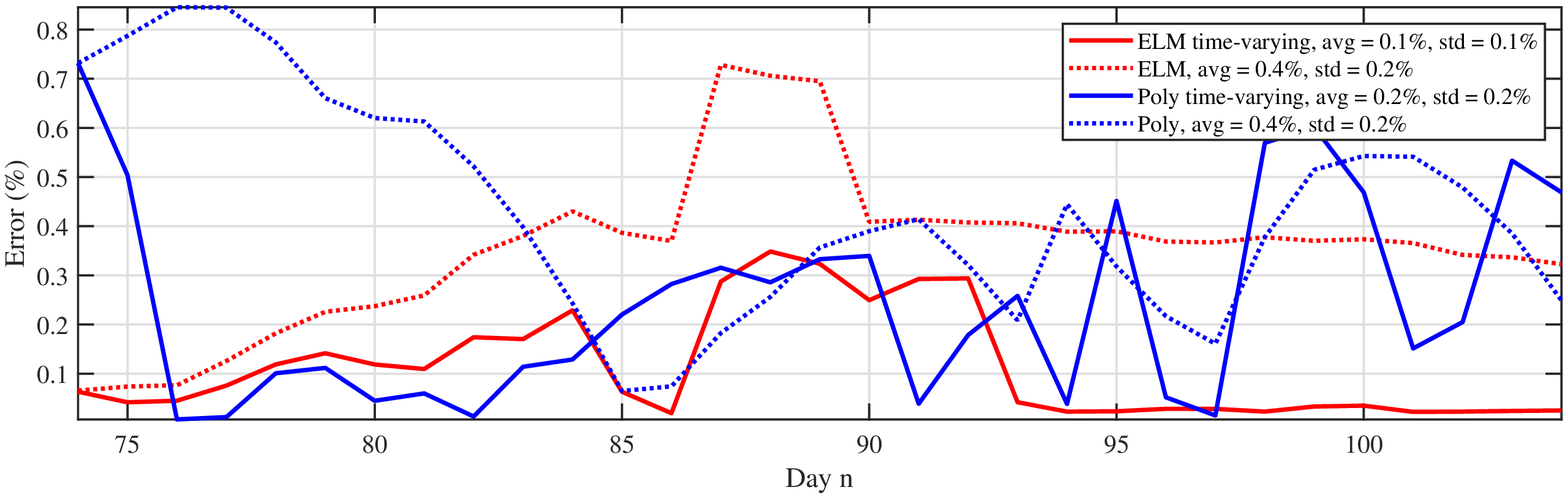} 
        \caption{China}
    \end{subfigure} \\
    \begin{subfigure}{\textwidth}
        \centering
        \includegraphics[width=\textwidth]{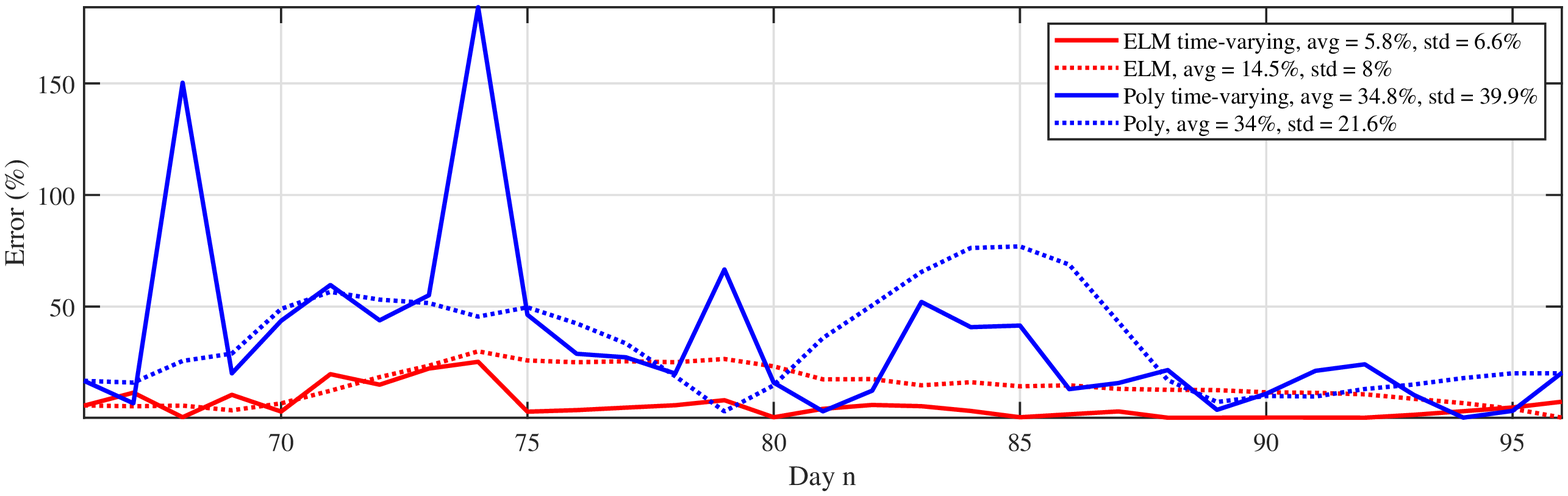} 
        \caption{India}
    \end{subfigure} \\
    \begin{subfigure}{\textwidth}
        \centering
        \includegraphics[width=\textwidth]{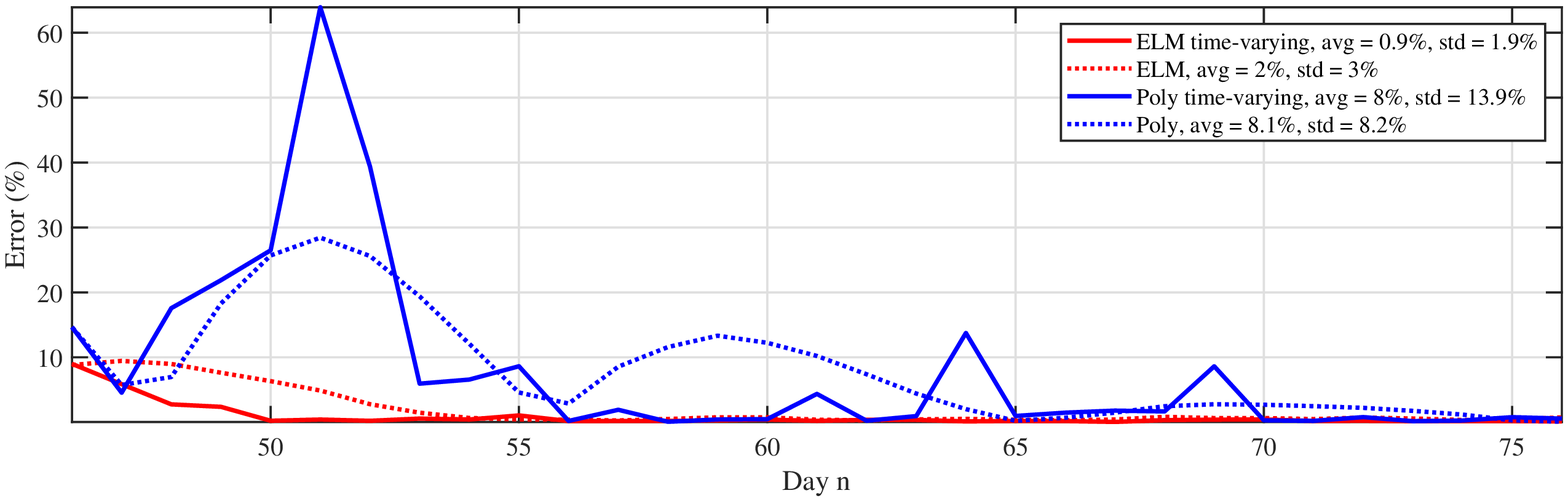}
        \caption{Iran}
    \end{subfigure} \\
    \begin{subfigure}{\textwidth}
        \centering
        \includegraphics[width=\textwidth]{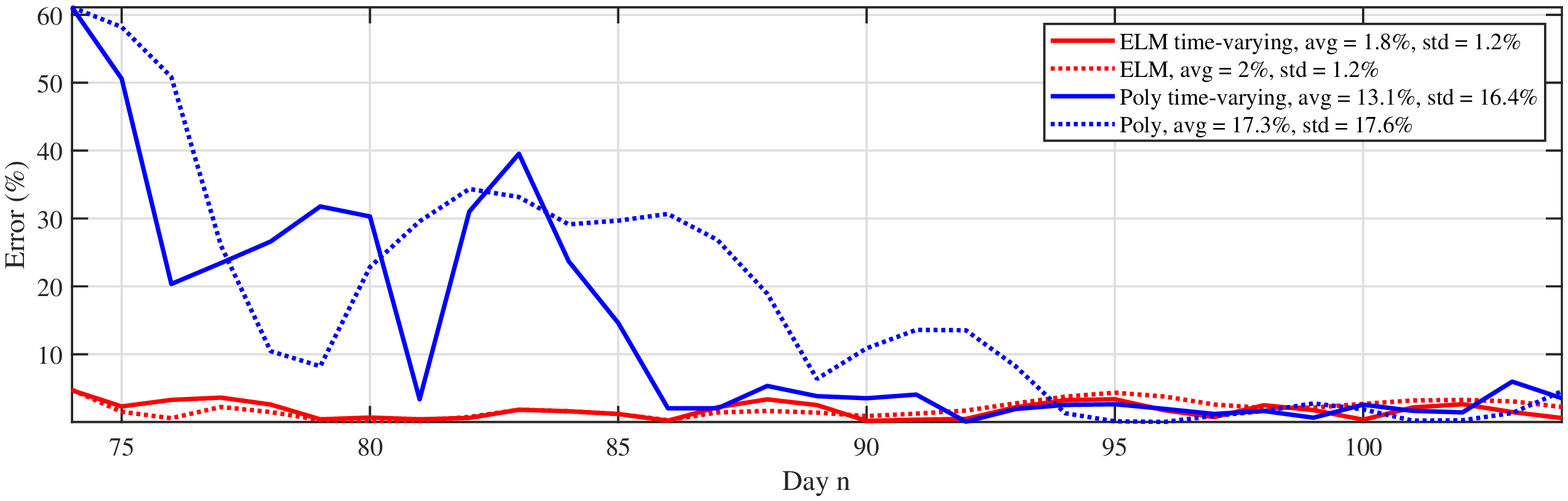} 
        \caption{USA}
    \end{subfigure} \\
    \caption{Daily error percentage of the last 31 days of 12 countries for ELM and polynomial regression. Here, $\tau=3$.}
    \label{fig:Rest_tau_3}
\end{figure*}

\begin{figure*}[t!]
    \begin{subfigure}{\textwidth}
        \centering
        \includegraphics[width=\textwidth]{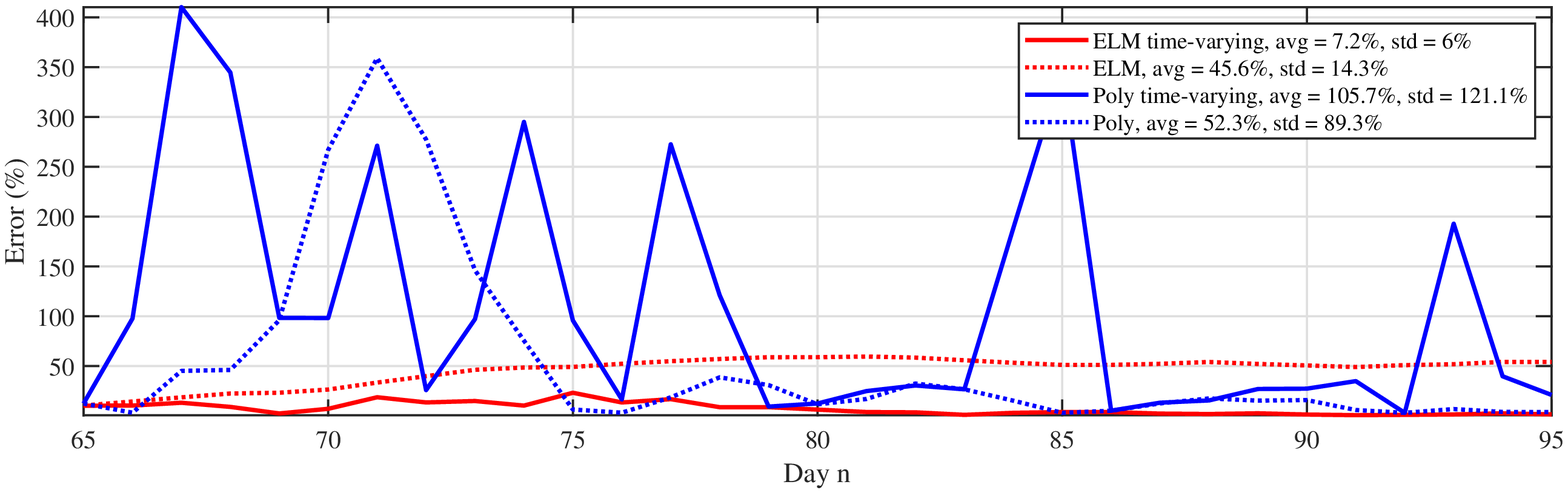}
        \caption{Sweden}
    \end{subfigure}    \\
    \begin{subfigure}{\textwidth}
        \centering
        \includegraphics[width=\textwidth]{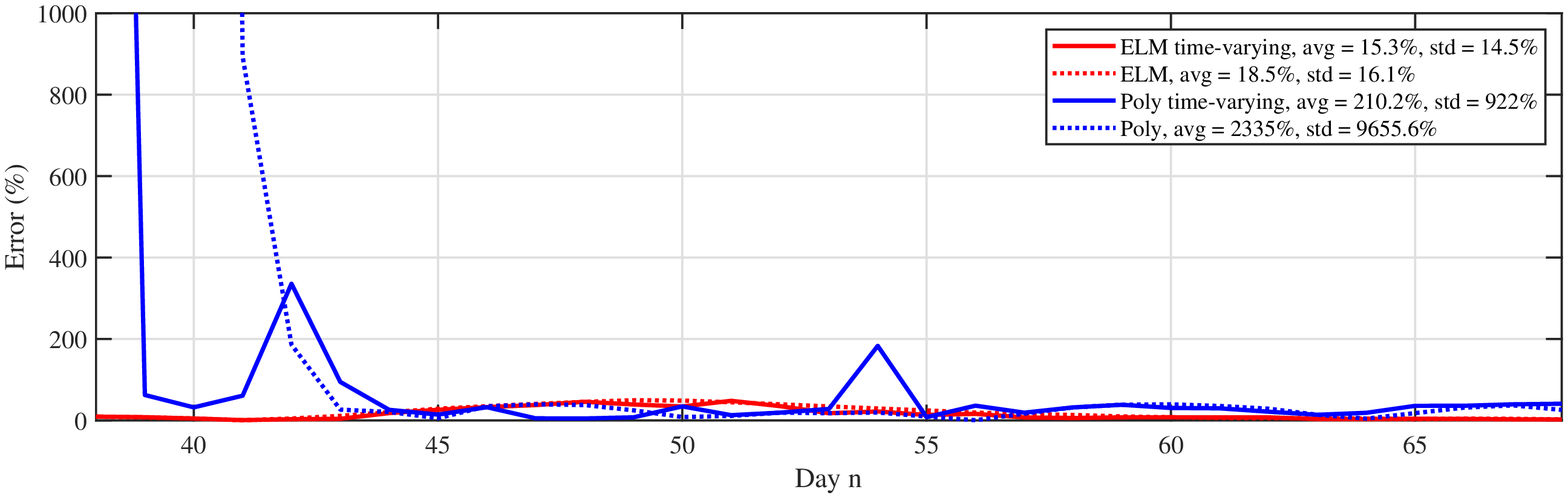}
        \caption{Denmark}
    \end{subfigure} \\
    \begin{subfigure}{\textwidth}
        \centering
        \includegraphics[width=\textwidth]{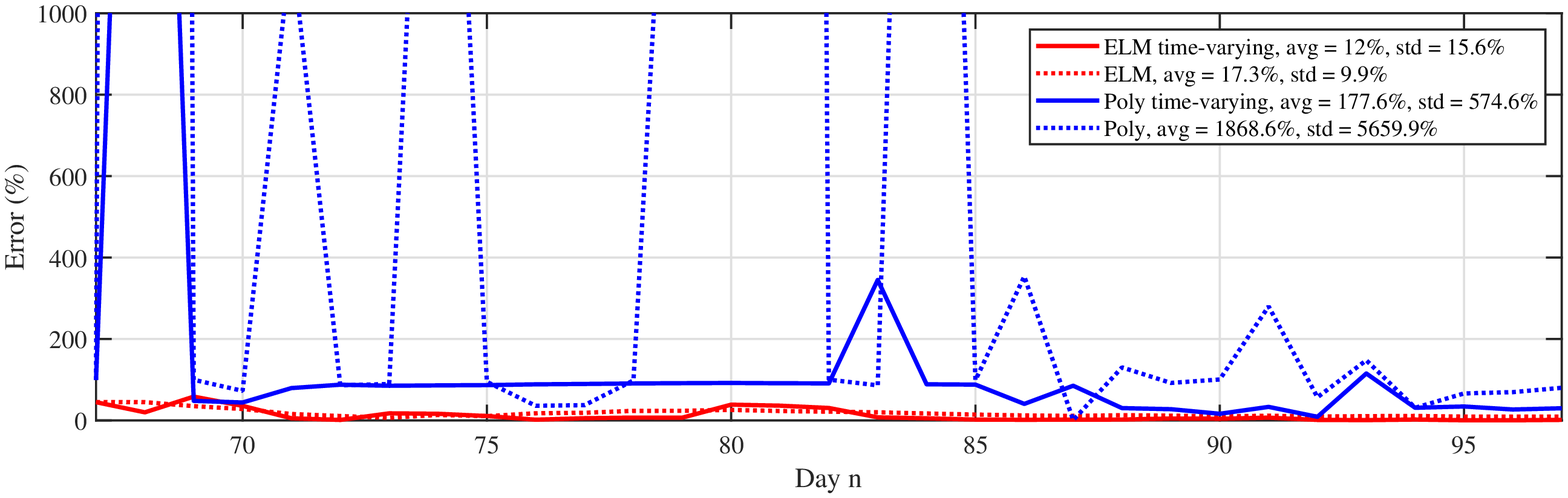}
        \caption{Finland}
    \end{subfigure}    \\
    \begin{subfigure}{\textwidth}
        \centering
        \includegraphics[width=\textwidth]{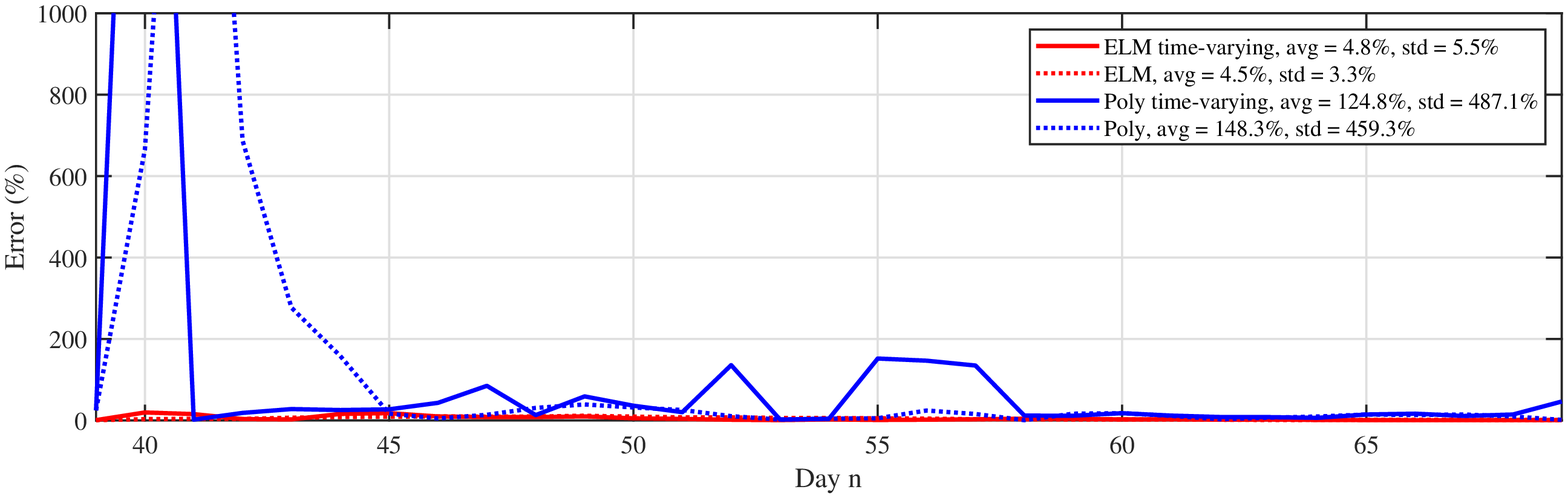} 
        \caption{Norway}
    \end{subfigure} \\
    \caption{Daily error percentage of the last 31 days of 12 countries for ELM and polynomial regression. Here, $\tau=7$.}
    \label{fig:Scandinavia_tau_7}
\end{figure*}

\begin{figure*}[t!]
    \begin{subfigure}{\textwidth}
        \centering
        \includegraphics[width=\textwidth]{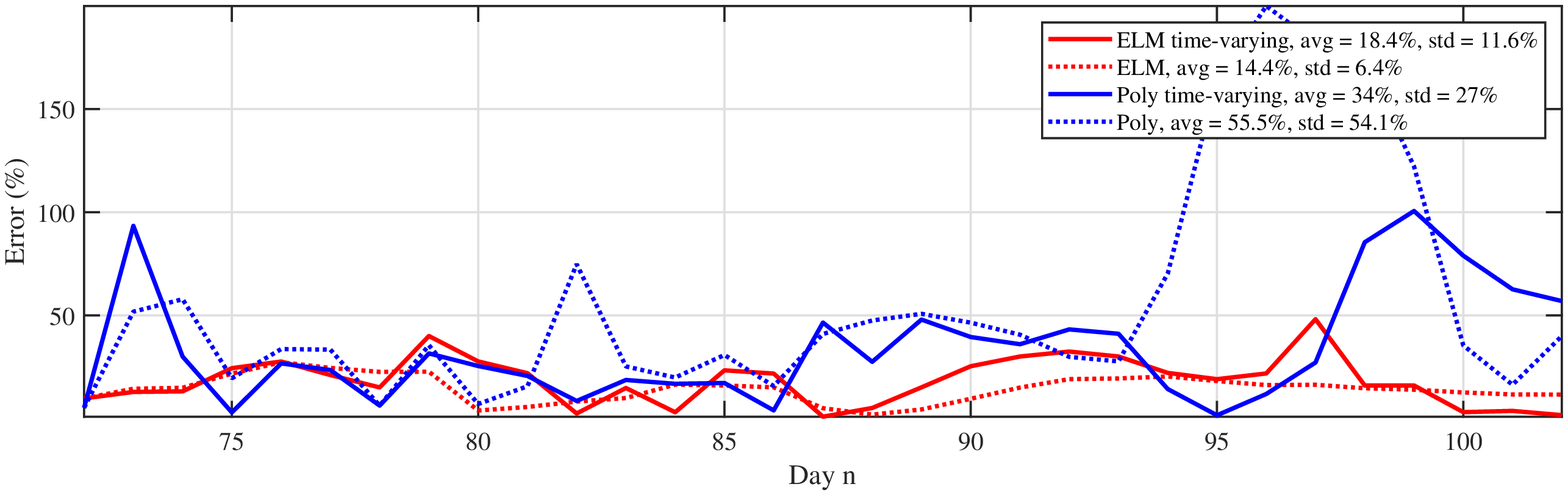} 
        \caption{France}
    \end{subfigure} \\
    \begin{subfigure}{\textwidth}
        \centering
        \includegraphics[width=\textwidth]{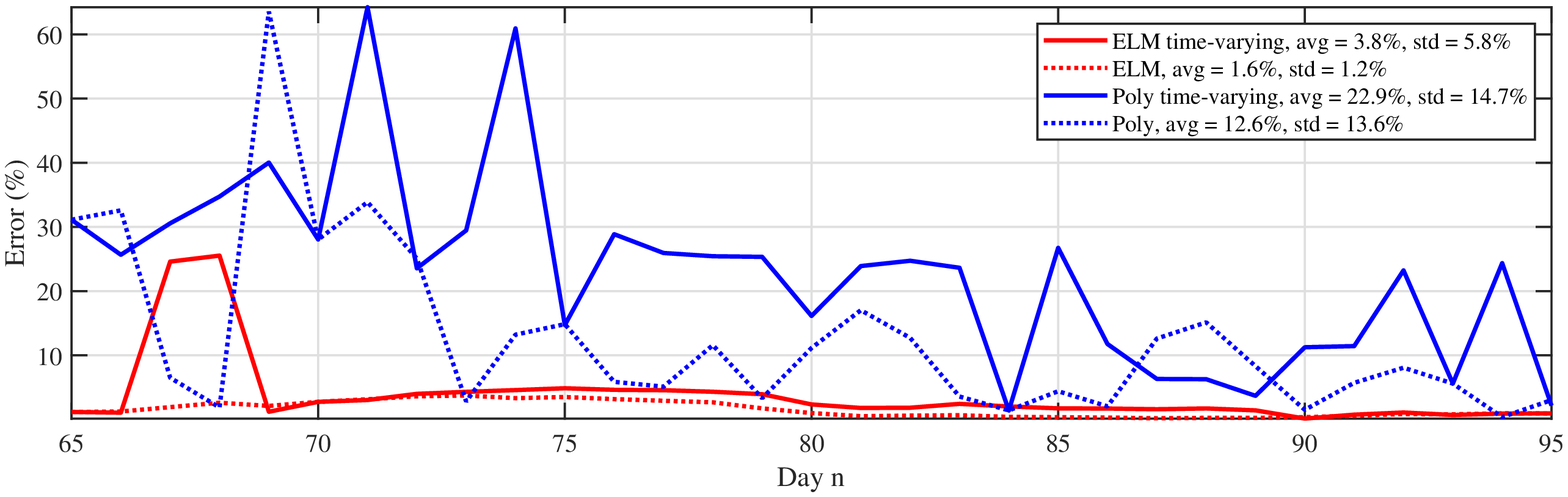}
        \caption{Italy}
    \end{subfigure}    \\
    \begin{subfigure}{\textwidth}
        \centering
        \includegraphics[width=\textwidth]{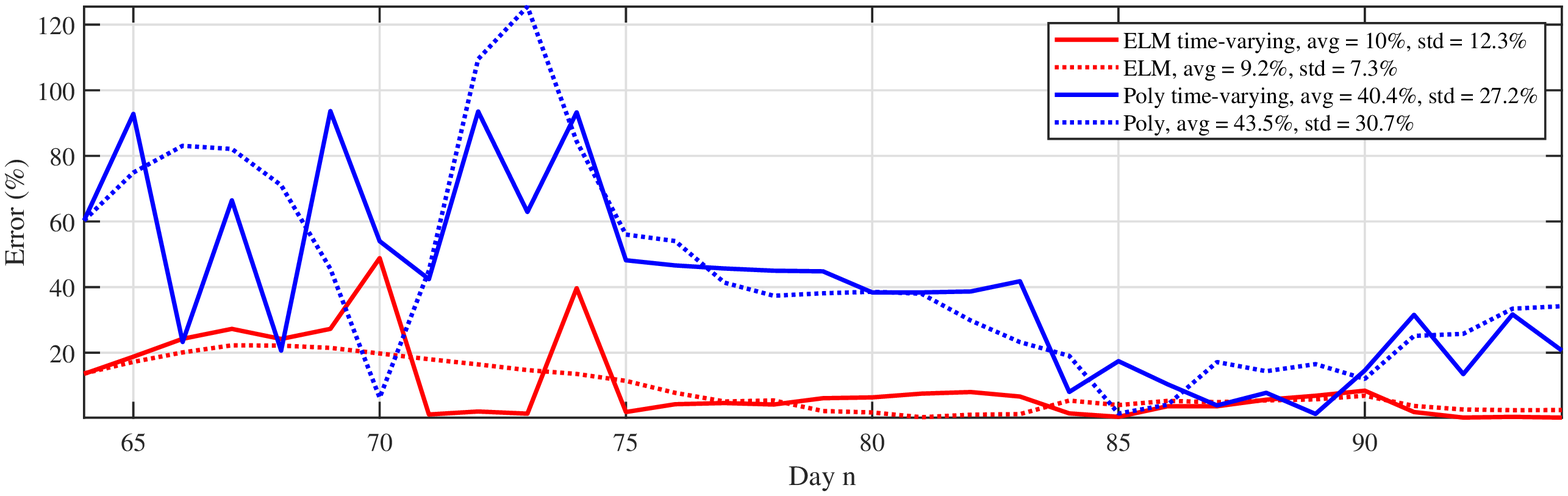}
        \caption{Spain}
    \end{subfigure} \\
    \begin{subfigure}{\textwidth}
        \centering
        \includegraphics[width=\textwidth]{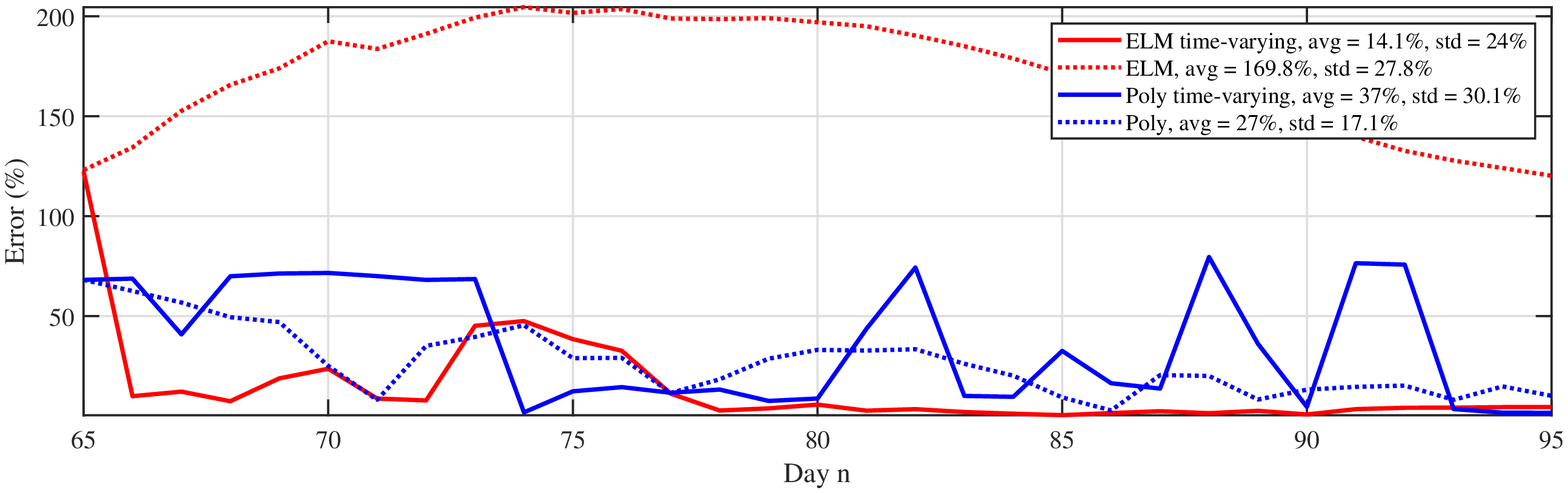} 
        \caption{UK}
    \end{subfigure} \\
    \caption{Daily error percentage of the last 31 days of 12 countries for ELM and polynomial regression. Here, $\tau=7$.}
    \label{fig:Europe_tau_7}
\end{figure*}

\begin{figure*}[t!]
    \begin{subfigure}{\textwidth}
        \centering
        \includegraphics[width=\textwidth]{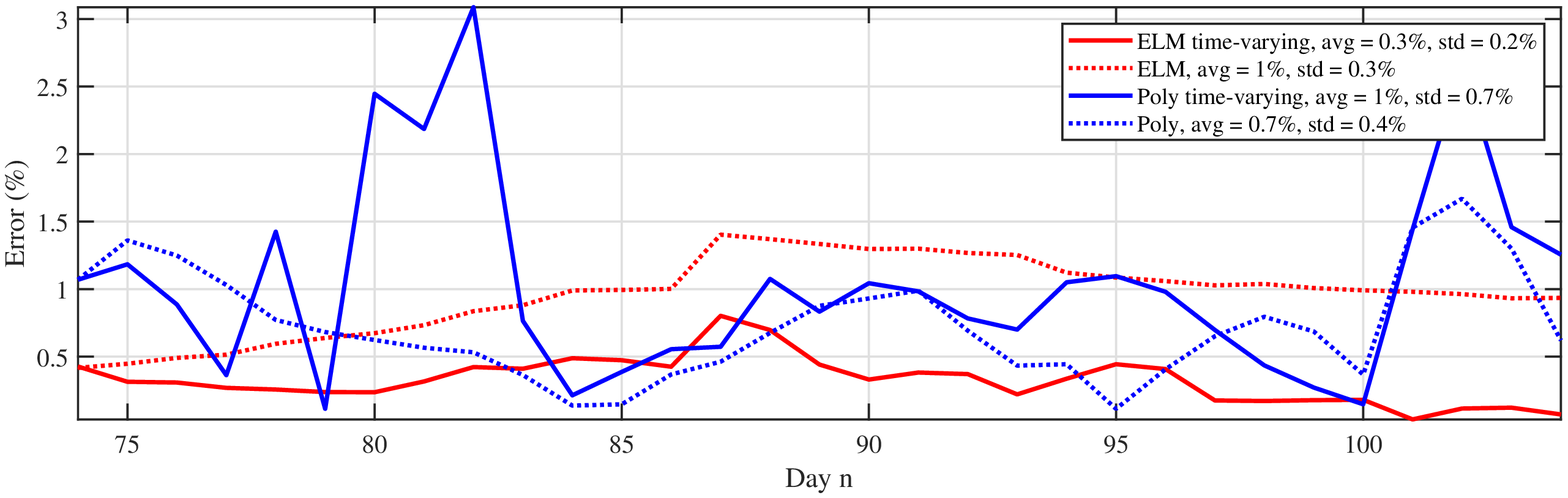} 
        \caption{China}
    \end{subfigure} \\
    \begin{subfigure}{\textwidth}
        \centering
        \includegraphics[width=\textwidth]{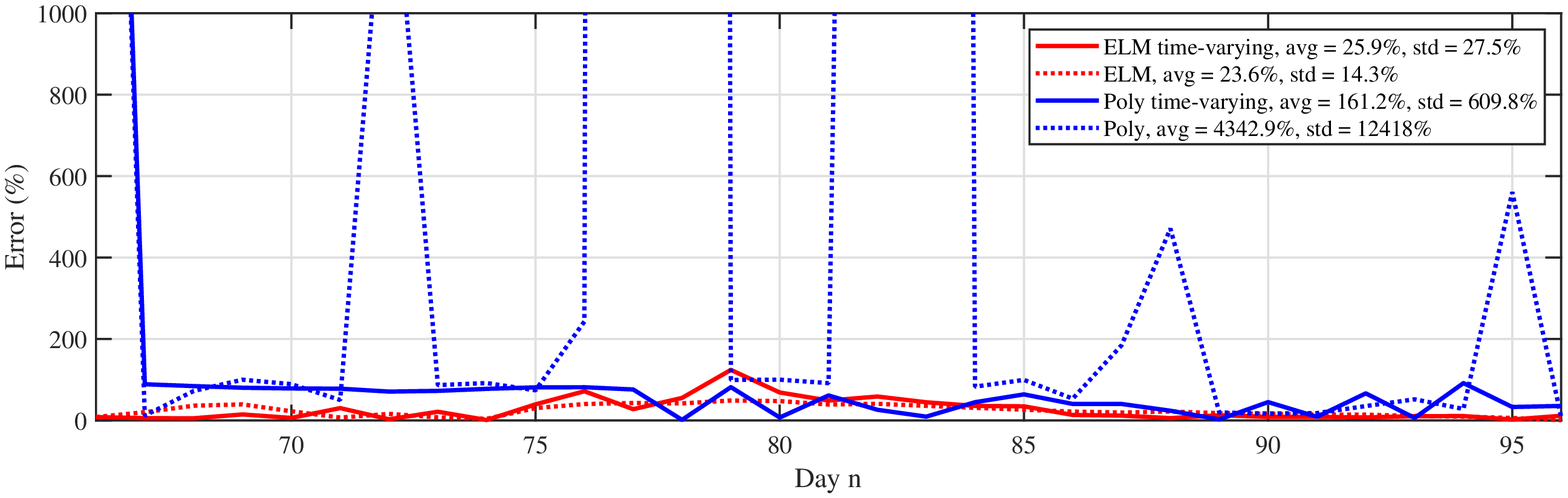} 
        \caption{India}
    \end{subfigure} \\
    \begin{subfigure}{\textwidth}
        \centering
        \includegraphics[width=\textwidth]{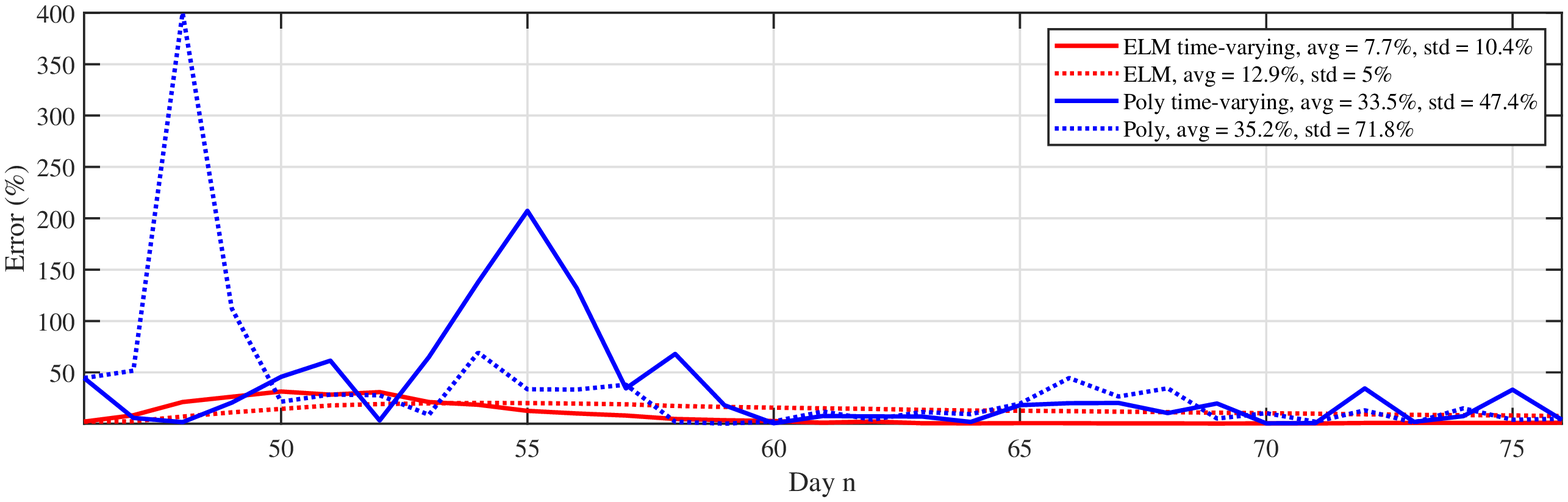}
        \caption{Iran}
    \end{subfigure} \\
    \begin{subfigure}{\textwidth}
        \centering
        \includegraphics[width=\textwidth]{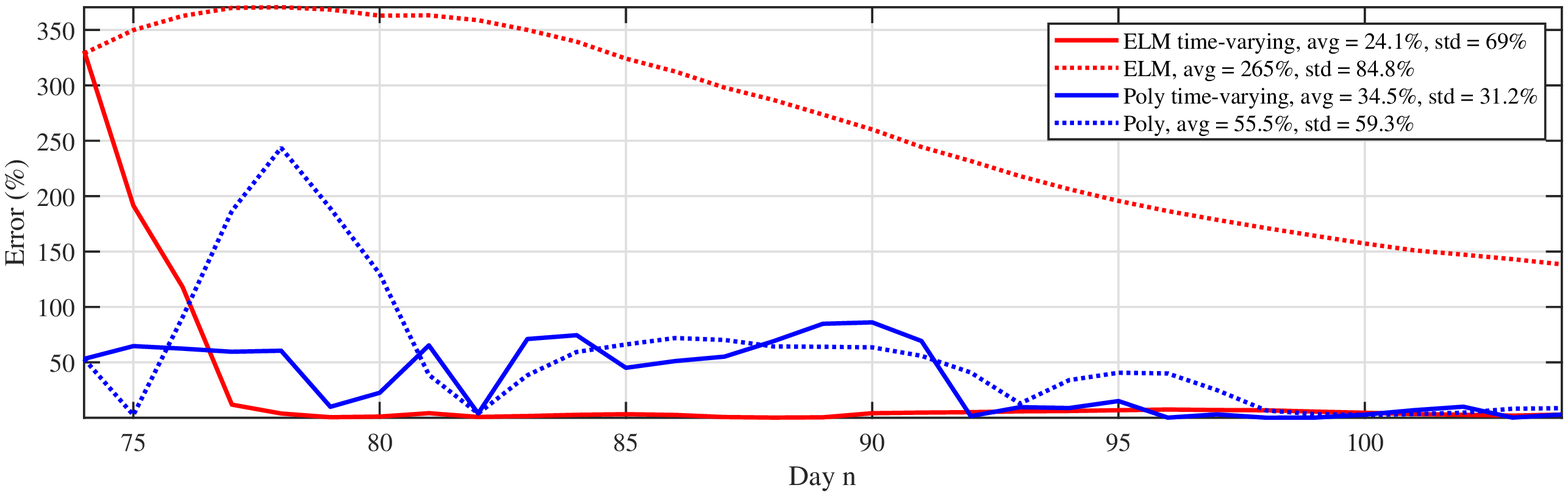} 
        \caption{USA}
    \end{subfigure} \\
    \caption{Daily error percentage of the last 31 days of 12 countries for ELM and polynomial regression. Here, $\tau=7$.}
    \label{fig:Rest_tau_7}
\end{figure*}

Histograms of error percentage of the four methods are shown in Figure \ref{fig:Histogram_tau} for different values of $\tau$. The histograms are calculated by using a nonparametric kernel-smoothing distribution over the past 31 days for all 12 countries. The daily error percentage for each country in Table \ref{table:Countries} is shown in Figures \ref{fig:Scandinavia_tau_1}-\ref{fig:Rest_tau_7}. Note that the reported error percentage of ELM is averaged over 100 Monte Carlo trials. 
The average and the standard deviation of the error over 31 days is reported (in percentage) in the legend of each of the figures for all four methods. It can be seen that daily cross-validation is crucial to preserve a consistent performance through-out the pandemic, resulting in a more accurate estimate. In other words, the variations of the time-series as $n$ increases is significant enough to change the statistics of the training and validation set, which, in turn, leads to different optimal hyperparameters as the length of the time-series grows. It can also be seen that ELM time-varying provides a more accurate estimate, especially for large values of $\tau$. Therefore, for the rest of the experiments, we only focus on ELM time-varying as our favored approach.

Another interesting observation is that the performance of ELM time-varying improves as $n$ increases. This observation verifies the general principle that neural networks typically perform better as more data becomes available. We report the average error percentage of ELM time-varying over the last 10 days of the time-series in Table \ref{table:Error_ELM_TV}. We see that as $\tau$ increases the estimation error increases. When $\tau = 7$, ELM time-varying works well for most of the countries. It does not perform well for France and India. This poor estimation for a few countries could be due to a significant amount of noise in the time-series data, even possibly caused by inaccurately reported daily cases.

\subsection{With the available data till May 12, 2020}
\label{subSec:May12}
In this subsection, we repeat the prediction based on the time-series data which is available until today May 12, 2020, for $\tau \in \{1,3,7\}$. In Subsection \ref{subSec:May4}, we predicted the total number of cases in Sweden on May 5, May 7, and May 11, 2020. The reported number of cases on these days for Sweden turned out to be 23216, 24623, and 26670, respectively, which is in the similar range of error that is reported in Table \ref{table:Error_ELM_TV}. 

\begin{figure*}[t!]
	\begin{subfigure}{\textwidth}
		\centering
		\includegraphics[width=\textwidth]{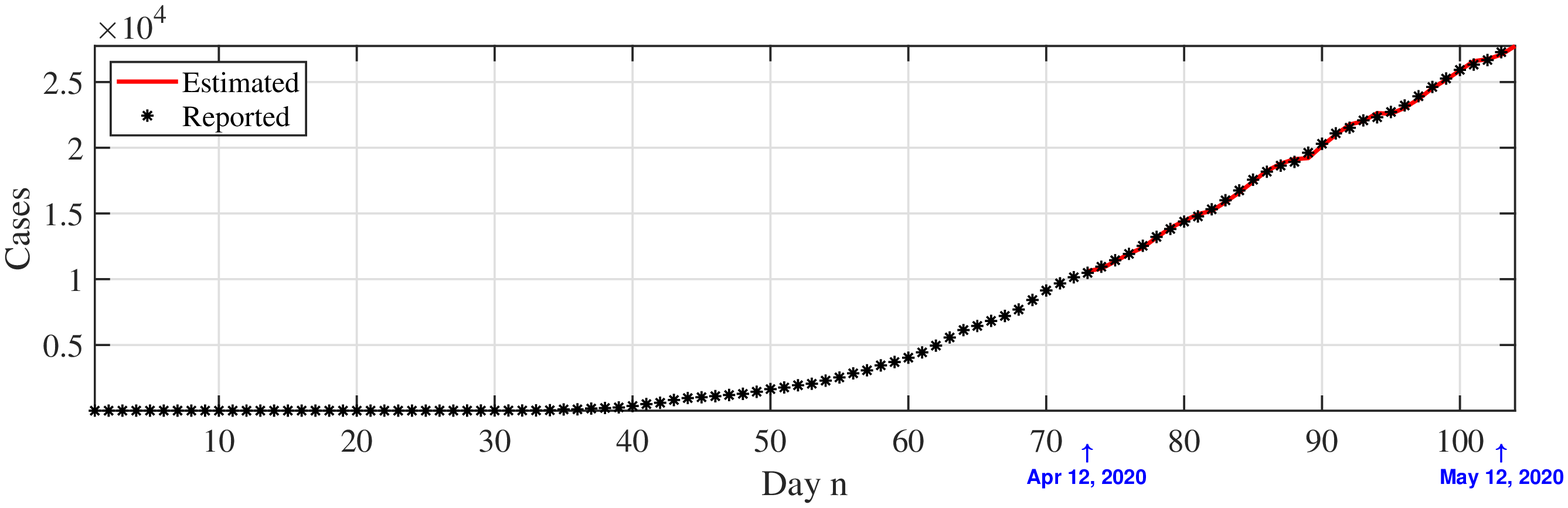}
		\caption{$\tau=1$}
	\end{subfigure} \\
	\begin{subfigure}{\textwidth}
		\centering
		\includegraphics[width=\textwidth]{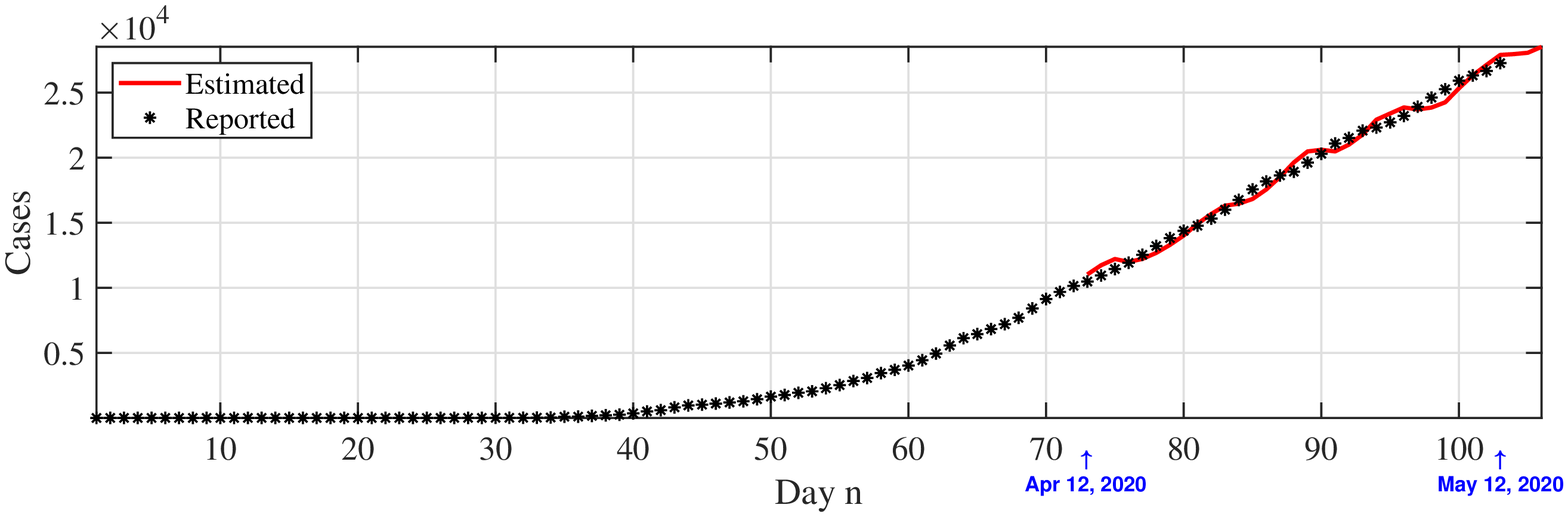}
		\caption{$\tau=3$}
	\end{subfigure} \\
	\begin{subfigure}{\textwidth}
		\centering
		\includegraphics[width=\textwidth]{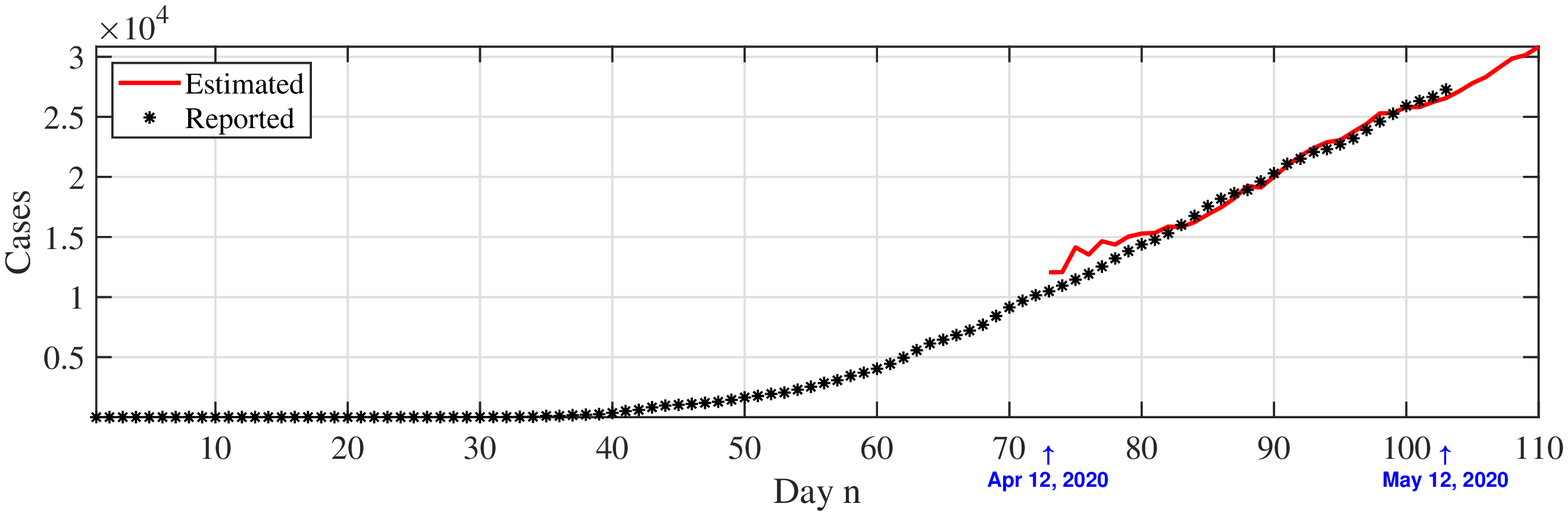}
		\caption{$\tau=7$}
	\end{subfigure} \\
	\caption{Reported and estimated cases after $\tau$ days over the last 31 days of Sweden for ELM time-varying. Update May 12.}
	\label{fig:Sweden_cases_May12}
\end{figure*}

We show the reported and estimated number of infection cases for Sweden by using ELM time-varying for different $\tau$'s in Figure \ref{fig:Sweden_cases_May12}. For each $\tau$, we estimate the number of cases up to $\tau$ days after which JHU data is collected. According to our experimental result, we predict that a total of 27737, 28522, and 30841 people will be infected in Sweden on May 13, May 15, and May 19, 2020, respectively.

\begin{figure*}[t!]
	\begin{subfigure}{\textwidth}
		\centering
		\includegraphics[width=\textwidth]{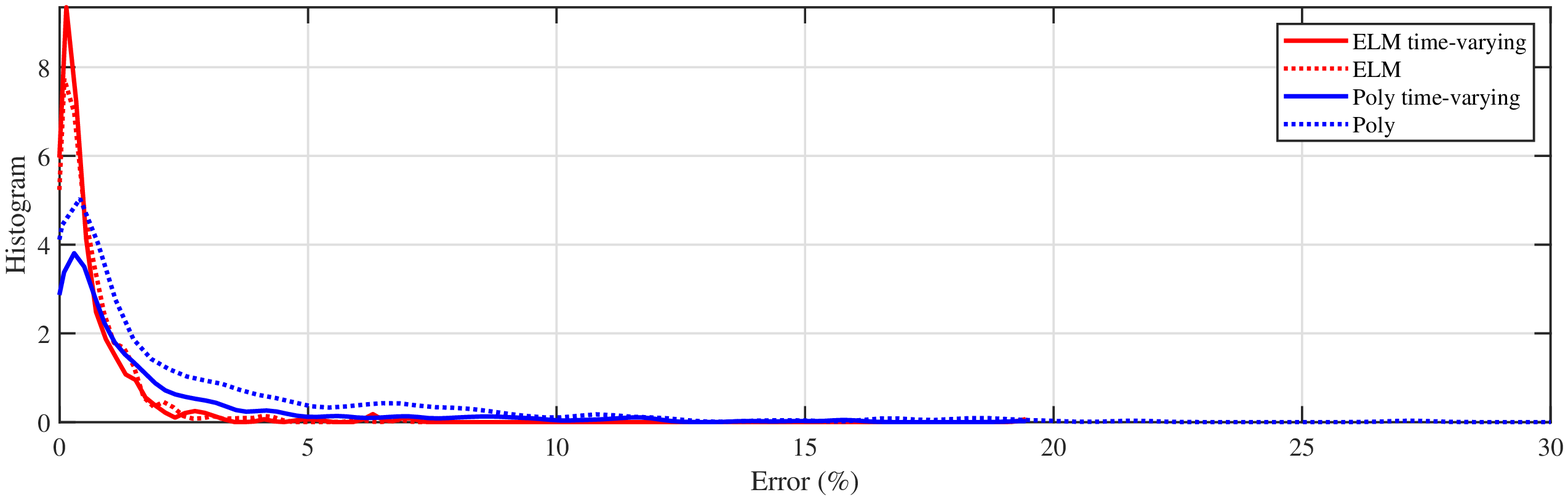}
		\caption{$\tau=1$}
		\vspace{15pt}
	\end{subfigure} \\
	\begin{subfigure}{\textwidth}
		\centering
		\includegraphics[width=\textwidth]{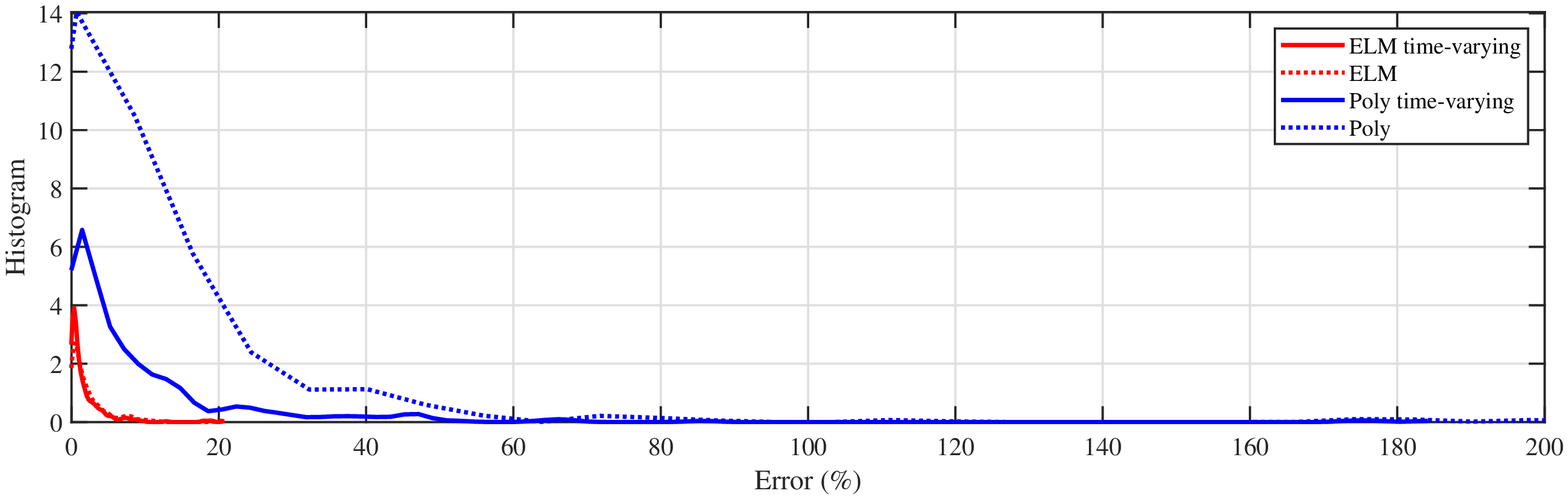}
		\caption{$\tau=3$}
		\vspace{15pt}
	\end{subfigure} \\
	\begin{subfigure}{\textwidth}
		\centering
		\includegraphics[width=\textwidth]{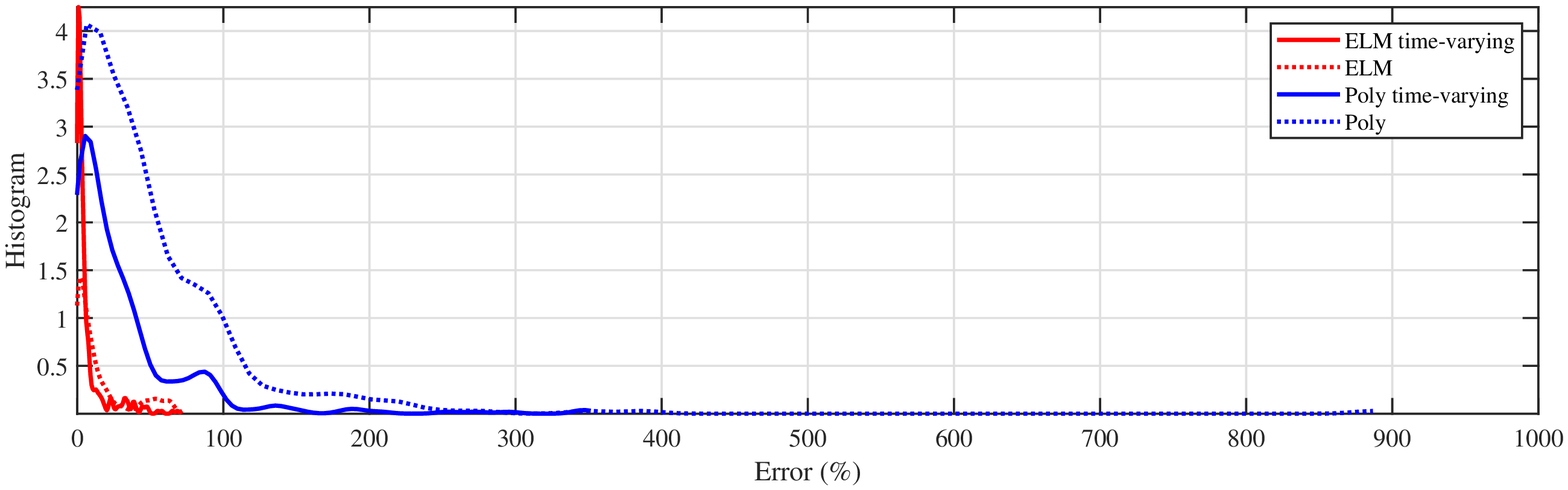}
		\caption{$\tau=7$}
		\vspace{15pt}
	\end{subfigure} \\
	\caption{Histogram of estimation error percentage over 31 days of all 12 countries for different values of $\tau$ for each of the ELM and polynomial approaches. Update May 12.}
	\label{fig:Histogram_tau_May12}
\end{figure*}

Histograms of error percentage of the four methods are shown in Figure \ref{fig:Histogram_tau_May12} for different values of $\tau$. These experiments verify that ELM time-varying is the most consistent approach as the length of the time-series increases from May 4 to May 12. We report the average error percentage of ELM time-varying over the last 10 days of the time-series in Table \ref{table:Error_ELM_TV_May12}. We see that as $\tau$ increases the estimation error increases. When $\tau = 7$, ELM time-varying works well for all of the countries except India, even though the number of training samples has increased compared to Subsection \ref{subSec:May4}.

\subsection{With the available data till May 20, 2020}
\label{subSec:May20}
In this subsection, we repeat the prediction based on the time-series data which is available until today May 20, 2020, for $\tau \in \{1,7,14\}$. In Subsection \ref{subSec:May12}, we predicted the total number of cases in Sweden on May 13, May 15, and May 19, 2020. The reported number of cases on these days for Sweden turned out to be 27909, 29207, and 30799, respectively, which is in the similar range of prediction error that is reported in Table \ref{table:Error_ELM_TV_May12}. 

\begin{figure*}[t!]
	\begin{subfigure}{\textwidth}
		\centering
		\includegraphics[width=\textwidth]{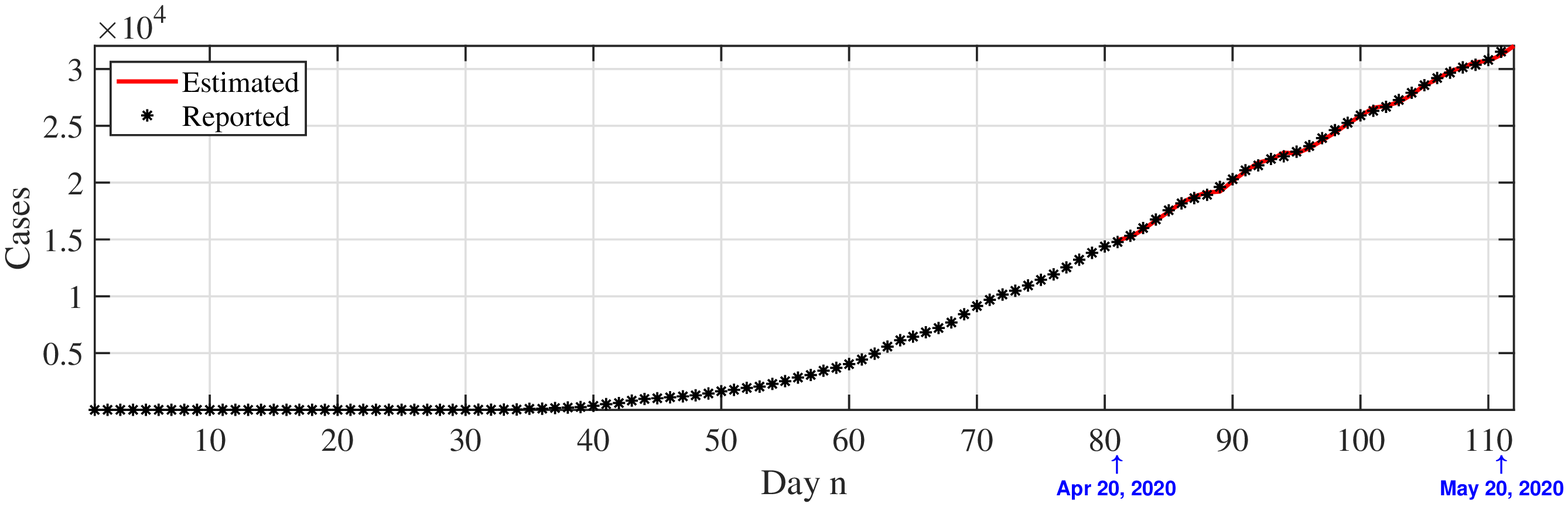}
		\caption{$\tau=1$}
	\end{subfigure} \\
	\begin{subfigure}{\textwidth}
		\centering
		\includegraphics[width=\textwidth]{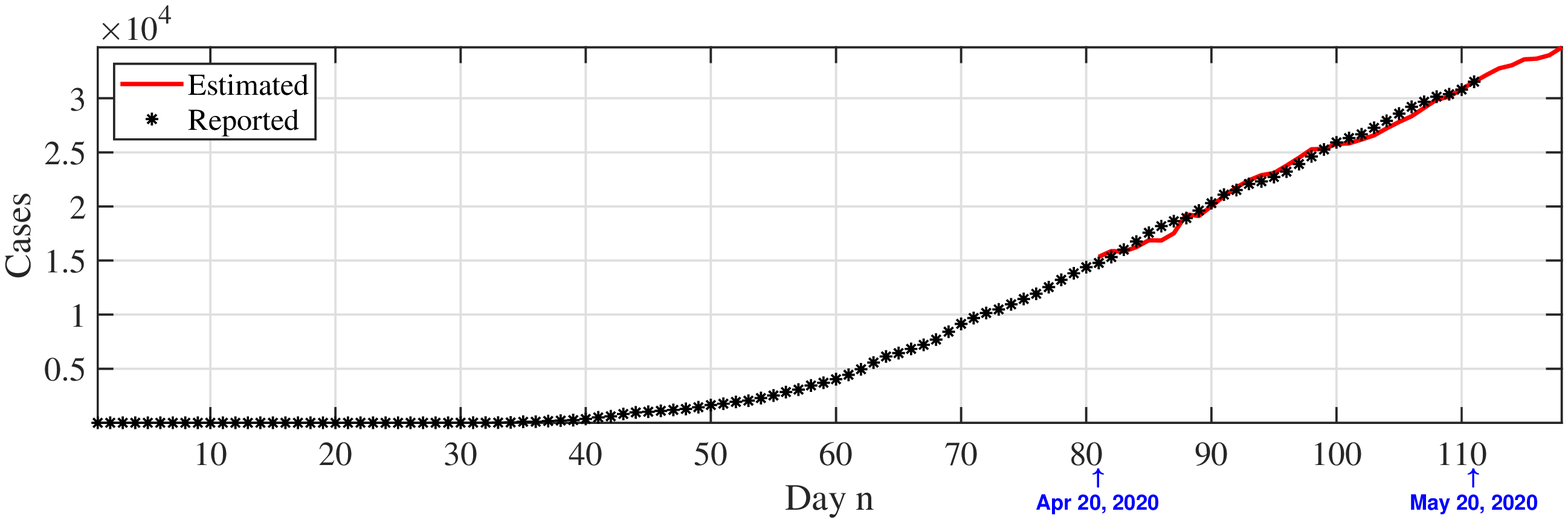}
		\caption{$\tau=7$}
	\end{subfigure} \\
	\begin{subfigure}{\textwidth}
		\centering
		\includegraphics[width=\textwidth]{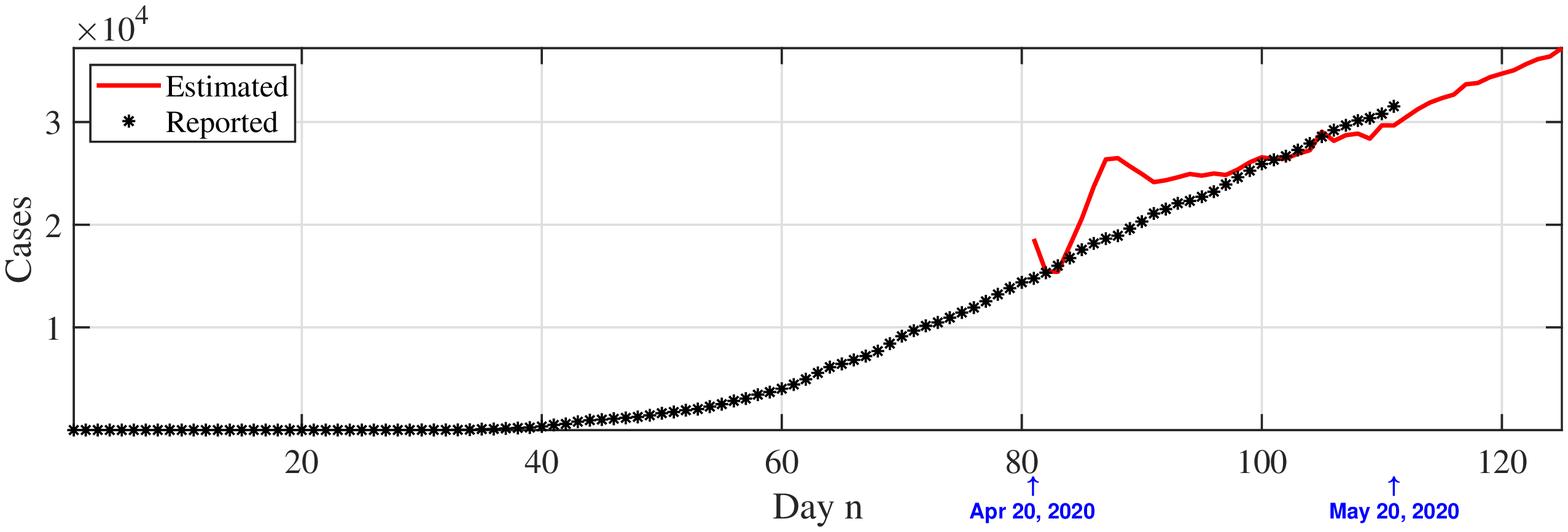}
		\caption{$\tau=14$}
	\end{subfigure} \\
	\caption{Reported and estimated cases after $\tau$ days over the last 31 days of Sweden for ELM time-varying. Update May 20.}
	\label{fig:Sweden_cases_May20}
\end{figure*}

We increase the prediction range $\tau$ in this subsection and we show the reported and estimated number of infection cases for Sweden by using ELM time-varying for $\tau=1,7,$ and $14$ in Figure \ref{fig:Sweden_cases_May20}. For each $\tau$, we estimate the number of cases up to $\tau$ days after which JHU data is collected. According to our experimental result, we predict that a total of 32032, 34702, and 37188 people will be infected in Sweden on May 21, May 27, and June 3, 2020, respectively.

\begin{figure*}[t!]
	\begin{subfigure}{\textwidth}
		\centering
		\includegraphics[width=\textwidth]{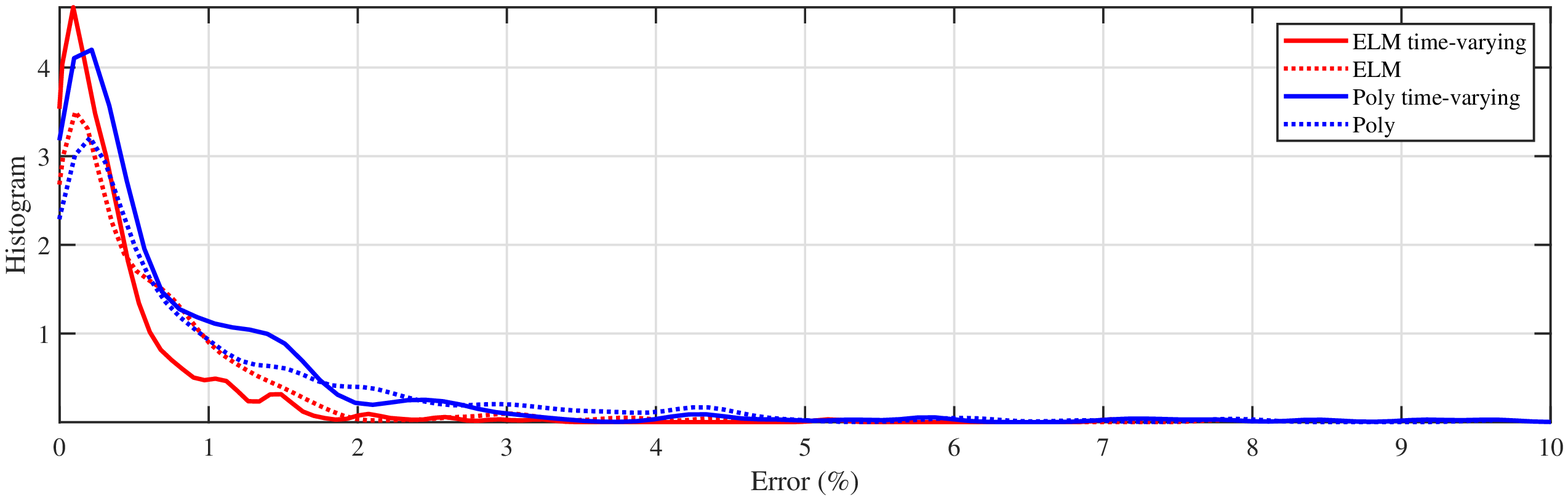}
		\caption{$\tau=1$}
		\vspace{15pt}
	\end{subfigure} \\
	\begin{subfigure}{\textwidth}
		\centering
		\includegraphics[width=\textwidth]{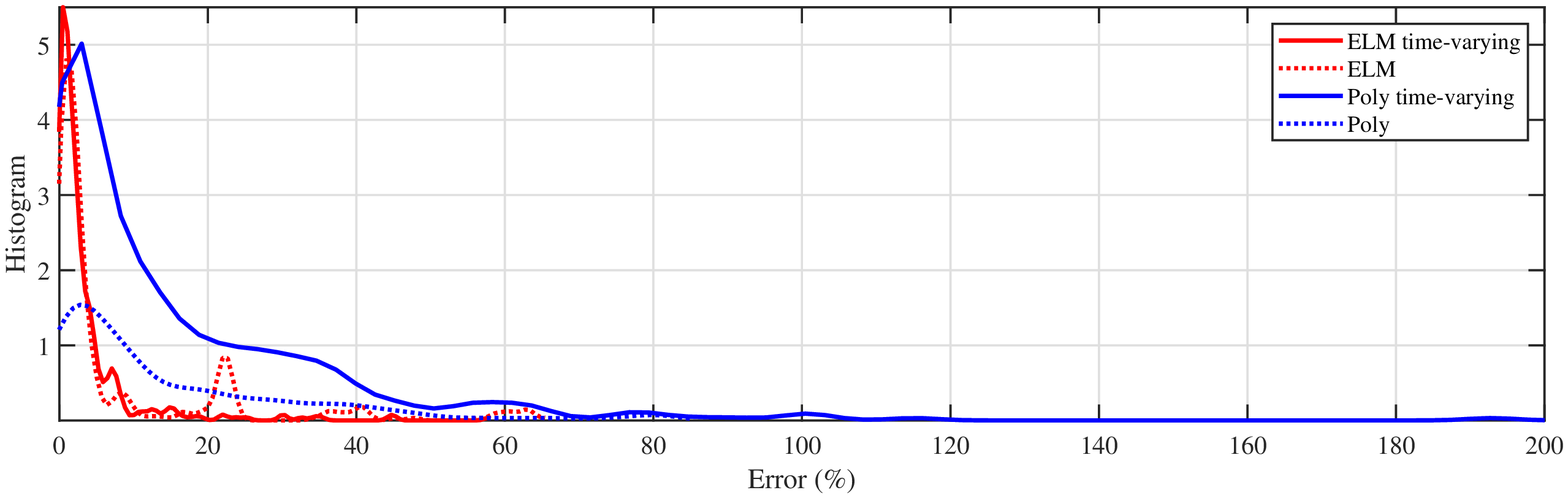}
		\caption{$\tau=7$}
		\vspace{15pt}
	\end{subfigure} \\
	\begin{subfigure}{\textwidth}
		\centering
		\includegraphics[width=\textwidth]{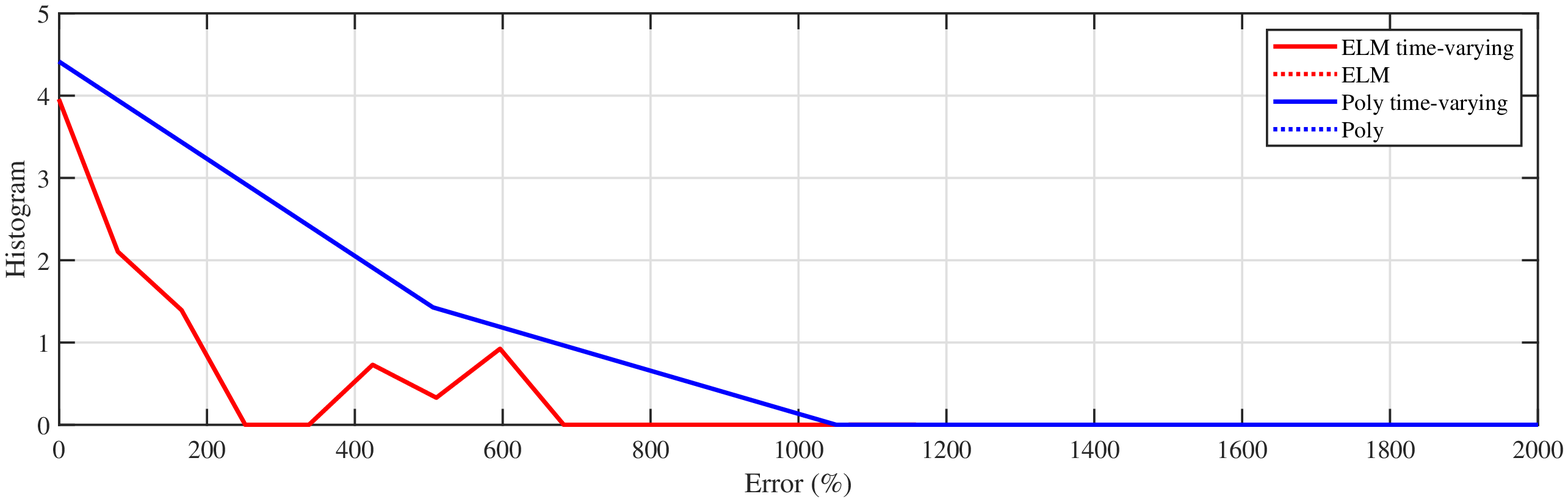}
		\caption{$\tau=14$}
		\vspace{15pt}
	\end{subfigure} \\
	\caption{Histogram of estimation error percentage over 31 days of all 12 countries for different values of $\tau$ for each of the ELM and polynomial approaches. Update May 20.}
	\label{fig:Histogram_tau_May20}
\end{figure*}

Histograms of error percentage of the four methods are shown in Figure \ref{fig:Histogram_tau_May20} for different values of $\tau$. These experiments verify that ELM time-varying is the most consistent approach as the length of the time-series increases from May 12 to May 20. We report the average error percentage of ELM time-varying over the last 10 days of the time-series in Table \ref{table:Error_ELM_TV_May20}. We see that as $\tau$ increases the estimation error increases. When $\tau = 7$, ELM time-varying works well for all of the countries so we increase the prediction range to 14 days. We observe that ELM time-varying fails to provide an accurate estimate for several countries such as France, India, Iran, and USA. This experiment shows that long-term prediction of the spread COVID-19 can be investigated as an open problem. However, by observing Tables \ref{table:Error_ELM_TV}-\ref{table:Error_ELM_TV_May20}, we expect that the performance of ELM time-varying to improve in the future as the number of training samples increases during the pandemic.

\section{Conclusion}
We studied the estimation capabilities of two well-known approaches to deal with the spread of the COVID-19 pandemic. We showed that a small-sized neural network such as ELM provides a more consistent estimation compared to polynomial regression counterpart. We found that a daily update of the model hyperparameters is of paramount importance to achieve a stable prediction performance. The proposed models currently use the only samples of the time-series data to predict the future number of cases. A potential future direction to improve the estimation accuracy is to incorporate constraints such as infectious disease spread model, non-pharmaceutical interventions, and authority policies \cite{folkhal}.

\begin{table*}[t!]
    \centering
    \caption{Average estimation error in percentage ($\%$) over the last 10 days for ELM time-varying. Update May 4.}
    \label{table:Error_ELM_TV}
    \setlength{\tabcolsep}{7pt}
    \renewcommand{\arraystretch}{1.7}
    \begin{tabular}{ |c|c|c|c|c|c|c|c|c|c|c|c|c| } 
        \hline
        Country & Sweden & Denmark & Finland & Norway & France & Italy & Spain & UK & China & India & Iran & USA \\ \hline \hline
        1 day prediction & \textbf{0.9} & 0.5 & 0.9 & 0.2 & 0.8 & 0.1 & 0.5 & 0.3 & 0 & 0.7 & 0.1 & 0.4 \\ 
        \hline
        3 days prediction & \textbf{2.6} & 0.7 & 0.7 & 0.6 & 2 & 0.3 & 2.5 & 1.3 & 0 & 2.1 & 0.2 & 1.7 \\ 
        \hline
        7 days prediction & \textbf{2} & 4.8 & 2.2 & 1.2 & 18.2 & 1.1 & 3.1 & 3 & 0.2 & 8.8 & 0.6 & 4.9 \\ 
        \hline
    \end{tabular}
\end{table*} 

\begin{table*}[t!]
	\centering
	\caption{Average estimation error in percentage ($\%$) over the last 10 days for ELM time-varying. Update May 12.}
	\label{table:Error_ELM_TV_May12}
	\setlength{\tabcolsep}{7pt}
	\renewcommand{\arraystretch}{1.7}
	\begin{tabular}{ |c|c|c|c|c|c|c|c|c|c|c|c|c| } 
		\hline
		Country & Sweden & Denmark & Finland & Norway & France & Italy & Spain & UK & China & India & Iran & USA \\ \hline \hline
		1 day prediction & \textbf{0.7} & 0.2 & 0.6 & 0.1 & 0.5 & 0.1 & 0.3 & 0.3 & 0 & 0.9 & 0.3 & 0.2 \\ 
		\hline
		3 days prediction & \textbf{2.3} & 0.5 & 1 & 0.4 & 1.4 & 0.3 & 0.4 & 1.3 & 0 & 3.4 & 0.7 & 0.5 \\ 
		\hline
		7 days prediction & \textbf{1.8} & 1.2 & 1.4 & 0.8 & 3.6 & 0.7 & 1.3 & 3.8 & 0 & 10.9 & 2.9 & 2.3 \\ 
		\hline
	\end{tabular}
\end{table*} 

\begin{table*}[t!]
	\centering
	\caption{Average estimation error in percentage ($\%$) over the last 10 days for ELM time-varying. Update May 20.}
	\label{table:Error_ELM_TV_May20}
	\setlength{\tabcolsep}{7pt}
	\renewcommand{\arraystretch}{1.7}
	\begin{tabular}{ |c|c|c|c|c|c|c|c|c|c|c|c|c| } 
		\hline
		Country & Sweden & Denmark & Finland & Norway & France & Italy & Spain & UK & China & India & Iran & USA \\ \hline \hline
		1 day prediction & \textbf{0.5} & 0.2 & 0.4 & 0.1 & 0.2 & 0.1 & 0.3 & 0.2 & 0 & 0.5 & 0.3 & 0.2 \\ 
		\hline
		7 days prediction & \textbf{1.6} & 1.4 & 1.4 & 0.3 & 1.4 & 0.3 & 1.4 & 1.2 & 0 & 2.9 & 3.6 & 1.2 \\ 
		\hline
		14 days prediction & \textbf{3.3} & 6 & 1.8 & 2.6 & 19.4 & 0.8 & 2.2 & 8.8 & 0.1 & 33.9 & 10.4 & 11.3 \\ 
		\hline
	\end{tabular}
\end{table*} 

\ifCLASSOPTIONcaptionsoff
  \newpage
\fi

\bibliographystyle{IEEEbib}
\bibliography{main}

\end{document}

%% file: Header.tex
\newsavebox\mybox

